\documentclass{article} % For LaTeX2e
\usepackage{iclr2024_conference,times}

\usepackage{hyperref}
\usepackage{url}
\usepackage{booktabs}
\usepackage{amsmath}
\usepackage{amssymb}
\usepackage{subcaption}
\usepackage{multirow}
\usepackage{wrapfig}
\usepackage{graphicx}
\usepackage[export]{adjustbox}
\usepackage{enumitem}
\usepackage{algpseudocode}
\newcommand{\cites}[1]{\citeauthor{#1}'s \citeyearpar{#1}}

\usepackage{xcolor}
\definecolor{myblue}{RGB}{31, 119, 180}
\newcommand{\blue}[0]{\color{myblue}\bf}
\definecolor{myorange}{RGB}{255, 127, 14}

\definecolor{mygreen}{RGB}{44, 160, 44}
\newcommand{\green}[0]{\color{mygreen}\bf}
\definecolor{myred}{RGB}{214, 39, 40}
\newcommand{\red}[0]{\color{myred}\bf}
\definecolor{mypurple}{RGB}{128, 0, 128}

\definecolor{bblue}{HTML}{1F77B4}

\newcommand{\original}[0]{\includegraphics[height=0.8em]{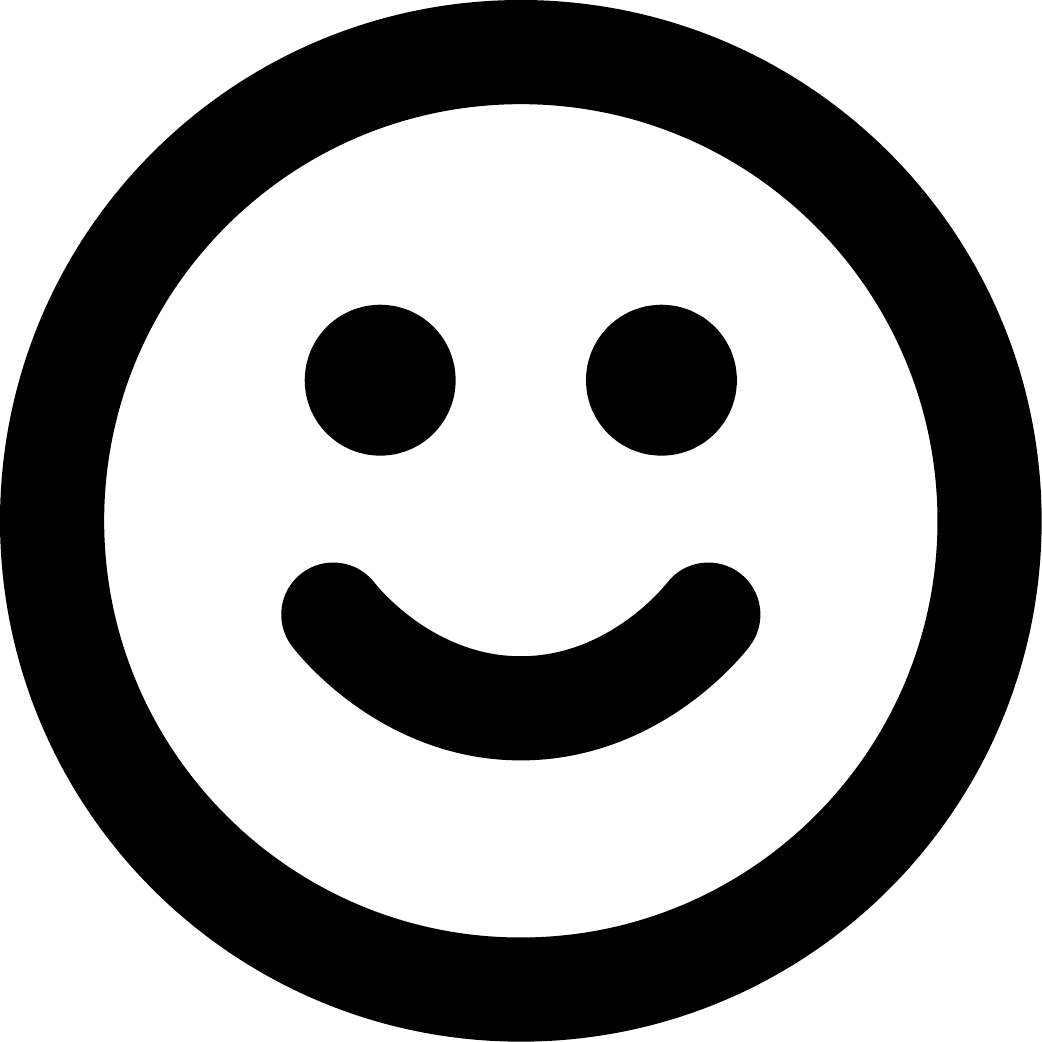}}
\newcommand{\patched}[0]{\includegraphics[height=0.8em]{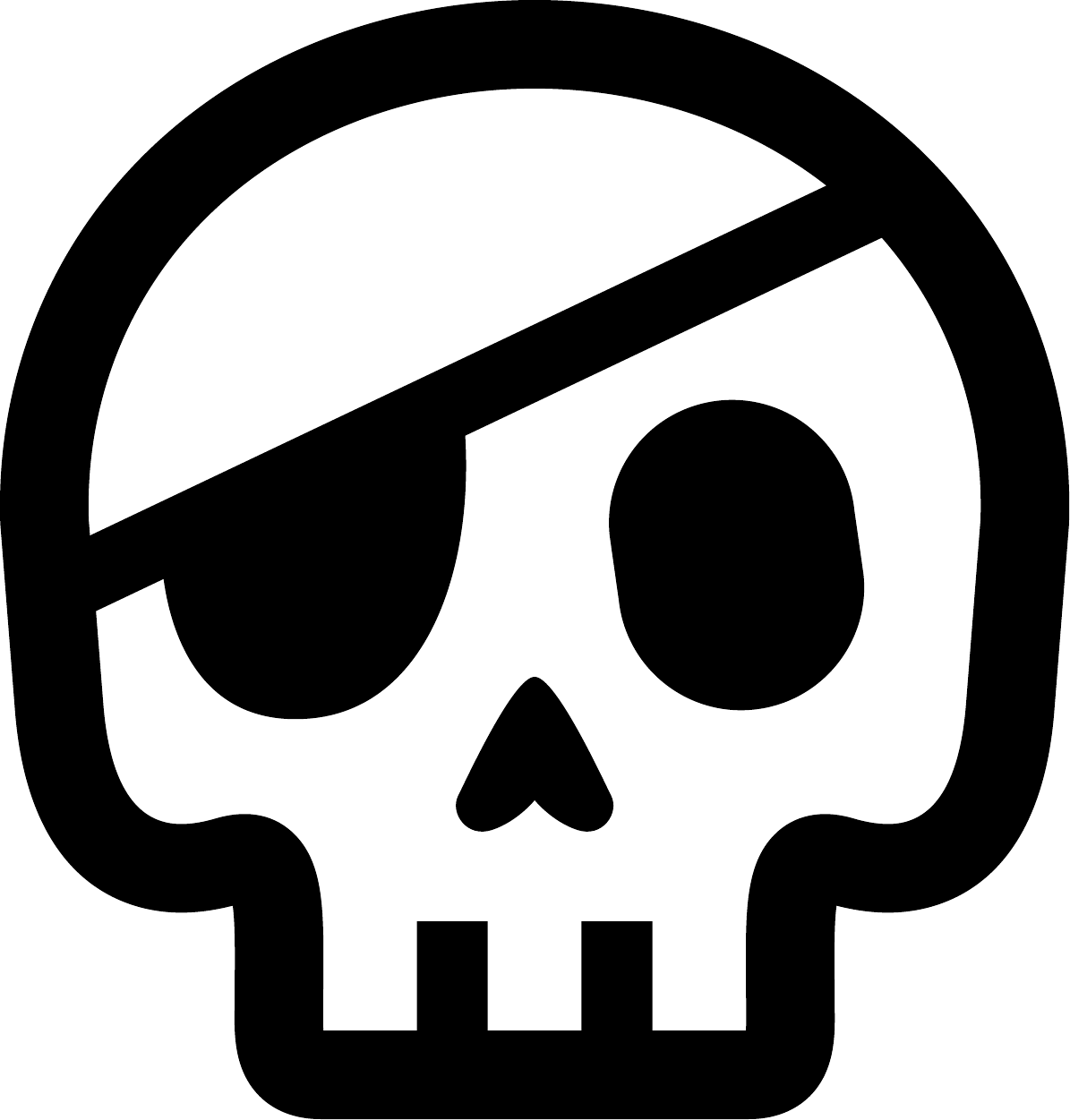}}

\usepackage{covington}

\title{What does the Knowledge Neuron Thesis Have to do with Knowledge?}

\author{%
Jingcheng Niu$^{14}$\\niu@cs.toronto.edu
\And
Andrew Liu$^2$\\a254liu@uwaterloo.ca
\And
Zining Zhu$^{134}$\\zzhu41@stevens.edu
\And
Gerald Penn$^{14}$\\gpenn@cs.toronto.edu
\AND
\vspace{-2.7em}\\
$^1$University of Toronto, $^2$University of Waterloo, $^3$Stevens Institute of Technology, $^4$Vector Institute}

\iclrfinalcopy % Uncomment for camera-ready version, but NOT for submission.
\begin{document}

\maketitle

\begin{abstract}
We reassess the Knowledge Neuron (KN) Thesis: an interpretation
of the mechanism underlying the ability of large language models to recall
facts from a training corpus. This nascent thesis proposes that facts are
recalled from the training corpus through the MLP weights in a manner resembling
key-value memory, implying in effect that ``knowledge'' is stored in the
network. Furthermore, by modifying the MLP modules, one can control the language
model's generation of factual information. The plausibility of the KN thesis has
been demonstrated by the success of KN-inspired model editing methods
\citep{daiKnowledgeNeuronsPretrained2022, mengLocatingEditingFactual2022}.

We find that this thesis is, at best, an oversimplification.
Not only have we found that we can edit the expression of certain linguistic
phenomena using the same model editing methods but, through a more comprehensive
evaluation, we have found that the KN thesis does not adequately explain the
process of factual expression. While it is possible to argue that the MLP
weights store complex patterns that are interpretable both syntactically and
semantically, these patterns do not constitute ``knowledge.'' To gain a more
comprehensive understanding of the knowledge representation process, we must
look beyond the MLP weights and explore recent models' complex layer structures
 and attention mechanisms.
\end{abstract}

\section{Introduction}

Recent research has highlighted the remarkable ability of large pretrained
language models (PLMs) to recall facts from a training corpus 
\citep{petroniLanguageModelsKnowledge2019}.  The underlying mechanism
by which this information is stored and retrieved within PLMs, however, remains a subject
of intensive investigation. The Knowledge Neuron (KN) Thesis has been recently
proposed as a novel framework for interpreting language models (LMs)
\citep{daiKnowledgeNeuronsPretrained2022, mengLocatingEditingFactual2022,
mengMassEditingMemoryTransformer2023}.
This thesis suggests that LMs operate akin to key-value memories, recalling
facts from the training corpus through the multi-layer perceptron (MLP)
weights.  Therefore, a significant implication of the KN thesis is that factual
information generation by LMs can be controlled by modifying the MLP modules. 
Should this manipulation of factual information recall become feasible, it could
lead to the development of language models that are more controllable, interpretable,
and factually aligned.

The plausibility of the KN thesis is demonstrated by the success of KN-inspired
model-editing methods.  \citet{daiKnowledgeNeuronsPretrained2022} argued that
relational facts can be localised to a handful of 2-5 MLP neurons.  They then
developed a method to identify these neurons using a search algorithm based on
an integral of gradients.  By manipulating the activation of these identified
neurons (\textit{KN edit}), they managed to alter the model's response to
fill-in-the-blank cloze tasks and generate counterfactual information without
additional fine-tuning.  In a parallel approach, 
\citet{mengLocatingEditingFactual2022} proposed a more intricate model wherein
factual recall occurs in two critical locations, each incorporating a different
module. In this model, the mid-layer MLP retrieves the fact, and an attention
module copies it into the output response at the topmost layer.  Despite this
proposed two-step process, their proposed model editing method, Rank-One Model
Editing (ROME), only modifies MLP weights, much as KN edit only modifies MLP
activations without editing attention modules.

\begin{figure}
\centering
\vspace{-1.5em}
\includegraphics[width=0.85\linewidth]{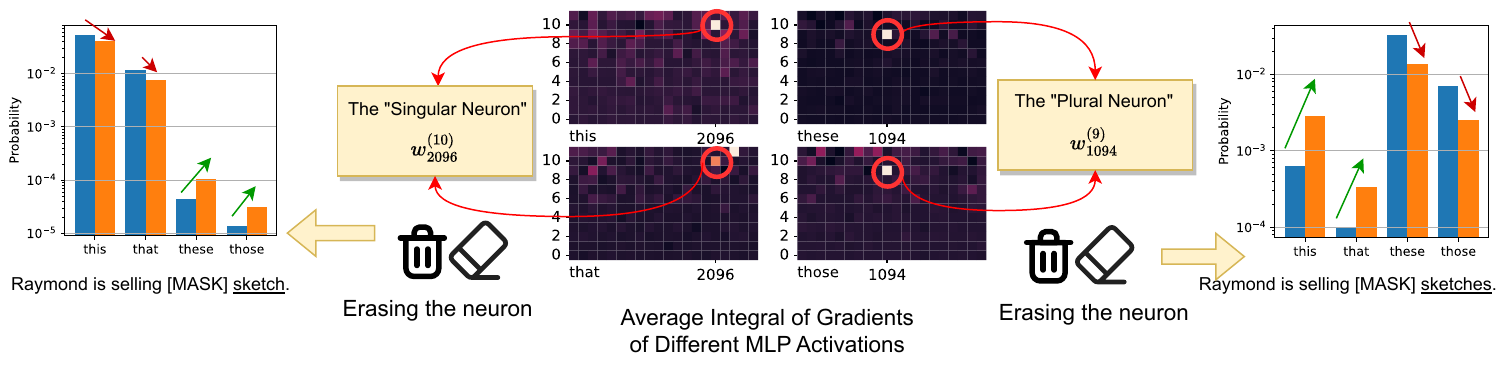}
\vspace{-0.7em}
\caption{Syntactic phenomena can be located and edited using existing model
editing methods.
The integrated gradient of singular determiner ({\it this}, {\it that}) and
plural determiner ({\it these}, {\it those}) form two distinct groups. 
Erasing these neurons leads to output probability changes.}
\label{fig:det_n_agr}
\vspace{-4em}
\end{figure}

While the efficacy of these model editing methods has been showcased in
simple fill-in-the-blank cloze tasks, the appraisal of such achievements
mainly rests on basic paraphrasing of the prompts, as outlined by  
\citet{yaoEditingLargeLanguage2023}, who introduced an additional
assessment metric, {\it portability}, finding that  model-editing methods
to date lack robustness.  Their performance is halved
when evaluated with the portability measure.  Building on this, we introduce
two new metrics.  First, a successful edit must demonstrate symmetry within
bijective relationships (e.g., with the assertion
{\it Ottawa is the capital of Canada}, the reciprocal {\it Canada's capital is
Ottawa} should also hold valid).  Second, a successful edit must extend to
synonym usage (e.g., {\it a dentist treats a toothache} and {\it a dentist treats
tooth pain} should be considered equivalent).  Our evaluation shows that existing 
model-editing methods are even less robust under these two new criteria.

It is practically impossible to exhaustively assess factual 
model-editing methods due to the difficulty in systematically dealing with
counterfactual
data.  The potential counterfactual replacements for Canada's capital are
seemingly endless.  Thus, beyond the introduction of the two new evaluation criteria
above, we
propose the evaluation of model-editing methods using syntactic constructions.
We have determined that the KN thesis applies just as reliably 
to syntactic phenomena (as illustrated in Figure \ref{fig:det_n_agr}).  Unlike
many facts, syntactic phenomena can provide rigorously defined targets for editing
through the use of so-called minimal pairs.  As a result, in this paper, we re-evaluate
the KN thesis by expanding the scope of our assessment to include more complex factual
patterns and syntactic phenomena.  This also speaks to a long-standing debate
regarding the formal {\it vs.} functional competence of language models 
\citep{mahowaldDissociatingLanguageThought2024} --- an LM's ability
to follow linguistic rules and patterns {\it vs.} its ability to apply
language in the real-world (see \S\ref{formal_vs_functional}). If we 
edit a model's expression of facts and linguistic phenomena using
the same approach, this could indicate that both the formal and functional
competencies of an LM are governed by the same underlying mechanisms.

Within the context of \cites{daiKnowledgeNeuronsPretrained2022} KN
framework, KN edit's efficacy is unsatisfactory.  Editing the KN
activations has only limited impact on categorical predictions.  The
effect of KN edit is only apparent in the shifts in the output
probability distributions of tokens.  The patterns that the method
localises also appeal to superficial cues such as word co-occurrence
frequencies.  We also find several critical shortcomings in the ROME
framework.  LMs process both linguistic and factual information in
phases, but the exact task distribution between the MLP and attention
modules appears to be more idiosyncratic than initially theorized
\citep{mengLocatingEditingFactual2022}.  ROME model editing only
superficially alters token association patterns, in a manner that is
inconsistent across the various expressions that may attend the same
underlying knowledge.  As a result, whatever is being manipulated
reflects none of the traditional tautologies that have been associated
with ``knowledge,'' as that term has been understood in philosophy
since the time of Aristotle.  When implemented on syntactic
constructions, furthermore, the influence of ROME's editing is limited
only to the word altered and no pivot that preserves any reasonable
standard of syntactic paraphrase, such as substitutability
\textit{salva veritate,} is forthcoming.  Furthermore, ROME fails
under our newly proposed symmetry and synonymy criteria.

We therefore argue for the position that the feed-forward MLP modules of the
transformer model do not store knowledge, but rather complex ``token expression
patterns.''  These token expression patterns can often be interpreted linguistically,
but the information that they express does not fit into linguistically or factually
defined categories.  A key-value, memory-based view of the language model is
overly simplistic in explaining the remarkable ability of recent PLM's formal,
and perhaps even functional, competence.  We need to investigate the rich
layer and attentive structure of recent PLMs more to arrive at a better understanding of 
their underlying mechanics.

In the following sections, we will first provide an overview of the KN thesis 
(\S\ref{sec:the_knowledge_neuron_thesis}).
Then we will evaluate two practices inspired by it: 
\cites{daiKnowledgeNeuronsPretrained2022} KN edit framework (\S\ref{sec:kn})
and \cites{mengLocatingEditingFactual2022} ROME framework (\S\ref{sec:rome}).
Finally, we will conclude the paper with a discussion 
(\S\ref{sec:discussion}).%
\footnote{The code, data and results are publicly available at \href{https://github.com/frankniujc/kn\_thesis}{https://github.com/frankniujc/kn\_thesis}.}

\section{The Knowledge Neuron Thesis}
\label{sec:the_knowledge_neuron_thesis}

\citet{gevaTransformerFeedForwardLayers2021} were among the first to propose
that the MLP modules in a transformer model behave like key-value memories. 
A typical MLP module in recent transformer-based PLMs has two layers.  They
argue that the first layer corresponds to keys, and the second layer, to values.%
\footnote{Despite the similarities in nomenclature, this is unrelated to the key
and value of self-attention.}
They found that each key neuron is triggered by human-interpretable shallow
input patterns such as periods of time that end with the letter ``{\it a}.'' 
Then, the corresponding value neurons distorted the next-token output
probability, until a final distribution is generated.

The KN thesis emerged as a result of this important discovery.
\citet{daiKnowledgeNeuronsPretrained2022} coined the term {\it knowledge neuron}
and ambitiously claimed that the keys and values within MLP modules not only
capture simple patterns but also store ``knowledge.''
They formulate an item of fact, such as
{\it \underline{Canada}'s capital is \underline{Ottawa}},
as a 3-tuple $(s,t,r)$, consisting of the source ($s$, {\it Canada}), the target
($t$, {\it Ottawa}) and the relation ($r$, {\it capital}) between them.
The authors asserted that this tuple can be localized to a small group of MLP
neurons typically found in the topmost layers of the language model, which they
identified by analysing the magnitude of the integrals of gradients among
prompts.  To support their claim, they conducted model-editing experiments.  By
suppressing the KNs (setting their activations to zero), they observed a
decrease in the probability of generating the correct original target ($t$),
while other tokens remained largely unaffected, demonstrating a ``minimally
invasive surgery.''
\citet{mengLocatingEditingFactual2022} proposed a refinement of 
\cites{daiKnowledgeNeuronsPretrained2022} model.  They employed a causal
mediation method \citep{finlaysonCausalAnalysisSyntactic2021} to form a more
intricate version of the KN thesis.  They argue that the factual association
process happens at two locations: a mid-layer MLP recalls the fact from
memory, and the topmost layer's attention model copies that information to the
final output.

There were similar investigations of neurons prior to the KN thesis.
\citet{durraniAnalyzingIndividualNeurons2020} observed the neurons of an
auxiliary probing model that was trained on BERT embeddings, not the neurons of
BERT itself.  Therefore, their analysis faced an all-too-common dilemma 
for probing: did they find insights about the language models or 
artefacts of the fine-tuning process \citep{hewittDesigningInterpretingProbes2019}?
\citet{finlaysonCausalAnalysisSyntactic2021} used causal
mediation analysis to study subject-verb agreement in GPT and XLNet 
\citep{yangXLNetGeneralizedAutoregressive2019}.  In particular, they observed a
difference in ratios between the verb with the correct inflection and one with the
incorrect inflection.  They then modify the prompt, see the probability
change and reason about the internal mechanisms of the model for expressing
subject-verb agreement.  They concluded that the upper-middle layers are more
relevant to the expression and that there are various levels of overlap between the
top 5\% neurons used to express agreement.  These insights, however, just as with
previous probing work, are still purely observational and largely preoccupied
with layers and network depth.  They are able to observe many characteristics of
the process, but still cannot cannot provide a satisfactory
understanding of how it happens.

More recently, there has been interest in utilizing large
language models (LLMs) to gain insight into the differing functionalities of individual
neurons.  Despite its title's strident claim that neurons in LMs
can be ``explained,'' \citet{billsLanguageModelsCan2023}
clarify that their model ``explains correlations,
not mechanisms.''  From a knowledge-representation standpoint,
their evaluation of LLM explanations is also entirely observational. 
When \citet{huangRigorouslyAssessingNatural2023a} reassessed the validity of these
explanations, even the most confident ones had
high error rates and little to no causal effects on the interventions that use the
explanations.  The LLM interpretation of LMs is still immature.

\subsection{Evaluating the KN Thesis: an Overview}
\label{sub:evaluating_the_kn_thesis_an_overview}

The effectiveness of a model-editing algorithm is customarily evaluated across
three dimensions \citep{yaoEditingLargeLanguage2023}:
(1) {\bf reliability}: whether the model can successfully change its output
from $t$ to $t^*$ (also referred to as an {\it efficacy score} by 
\citet{mengLocatingEditingFactual2022});
(2) {\bf generality}: whether the effect is applicable to rephrased
relations; and, (3) {\bf locality}: whether the edit impacts unrelated
relations.
\citet{yaoEditingLargeLanguage2023} stress, however, that the assessment of
generality is often constrained to simple paraphrasing.  This is typically
done by developing multiple templates for a specific relation.  For instance,
the relation {\it capital} can be structured as both
``The capital of [s] is [t].'' and  ``[s]'s capital is [t].''
Previous evaluations \citep{elazarMeasuringImprovingConsistency2021,
mengLocatingEditingFactual2022, mengMassEditingMemoryTransformer2023}
prematurely announced success when a model, edited on a first template, 
could be generalized to a second template.  Thus, 
\citet{yaoEditingLargeLanguage2023} recommended extending the assessment
parameters by introducing the concept of {\it portability}.
For example, having changed Watts
Humphrey's {\it alma mater} from Trinity College to Harvard University, the
model should return Boston instead of Dublin when asked about the city where
Watts Humphrey received his university education.  It was apparent that
model-editing methods present a markedly lower level of portability than
generality (50\% versus 90\%).
The evaluation of portability, on the other hand, requires new data annotation, which
can be costly.

Extending \citet{yaoEditingLargeLanguage2023}, we attempt a more comprehensive
evaluation of model editing of factual association with two extra criteria:
bijective symmetry and synonymous invariance.  Bijective symmetry
does not require new data collection and we can obtain data automatically from
previous corpora.
For a bijection relation such as {\it capital} or {\it capital of}, we should
see the model generalise $(s,t\rightarrow t^*,r)$ to $(t^*,s,r^{-1})$.
For example, if we change the capital of Canada to Rome, then the model should
also agree that Rome is the capital of Canada.  Similarly, an effective edit
should also be able to generalise across
synonyms.  If the model knows that a dentist treats toothaches, it should also
know that they also treat tooth pain.  Prior work 
\citep{elazarMeasuringImprovingConsistency2021} only used synonym replacement
on rephrasing the relation prompts --- we extend it to the source and the
target.

Several others have already questioned the validity
of the KN thesis. \citet{haseDoesLocalizationInform2023} identified discrepancies
between the results of causal tracing and the effects of ROME editing.
They concluded that a mechanistic understanding reveals insights on
the consequences of model editing.  To the
best of our knowledge, we are the first to comprehensively evaluate the KN
thesis using rigorously defined syntactic phenomena.  We consider three:
determiner-noun agreement, subject-verb agreement, and gender and number
agreement across anaphoric chains.

\subsection{Evaluating the KN Thesis on Syntactic Phenomena}
\label{sub:evaluating_the_kn_thesis_on_linguistic_phenomena}

Edit pairs for syntactic phenomena, by contrast, can be systematically extracted through the
formation of ``minimal pairs.''  For a grammatical sentence that expresses a
linguistic phenomenon, we can construct an ungrammatical sentence that minimally
differs from the original sentence in respect of one feature of grammatical acceptability.
For example, the phrase {\it this
student} can be changed to the ungrammatical counterpart, *{\it this
students}.  The BLiMP corpus \citep{warstadtBLiMPBenchmarkLinguistic2020} is one
of the most comprehensive and extensively utilised collections of such minimal
pairs.

We therefore propose to systematically evaluate the effect of model-editing methods
using syntactically differentiated prompts.  We define a similar 3-tuple $(s,t,p)$
that contains the source ($s$), the target ($t$) and the syntactic phenomenon ($p$).
Take the phenomenon determiner-noun agreement as an example.
In a grammatical sample sentence from a minimal pair, $s$ is the tokens that are
condition the expression of the target (the determiner), and $t$
is the tokens that differ within the pair (the noun).  The ungrammatical target
$t^*$, is the noun in the opposite form.  We then intervene with model editing,
and observe whether the model assigns a higher probability to $t$ than $t^*$.

\subsection{Editing Syntactic Phenomena \& the ``Formal vs Functional'' 
Distinction}
\label{formal_vs_functional}

If we can successfully edit facts as well as syntactic phenomena using the
same model-editing methods to the same degree, then it stands to reason that
the model follows a unified underlying mechanism for both factual and
syntactic information.  Choosing the correct city ({\it the Space
Needle is in Seattle/*Rome}) would be no different than choosing the correct 
verb form ({\it the apple is/*are red}).

\citet{mahowaldDissociatingLanguageThought2024} refers to a distinction between
the formal and functional competence of a language model: formal means
``knowledge of linguistic rules and patterns,'' and functional refers
to ``understanding and using language in the world.''  Syntactic
phenomena pertain to formal competence, and facts pertain to
functional competence, respectively.  NLP researchers sometimes informally
use the terms syntax and semantics to refer to this distinction.
BLiMP even refers to anaphoric gender agreement as morphological.
\citet{jawaharWhatDoesBERT2019} and \citet{tenneyBERTRediscoversClassical2019} believe that
syntactic information is located in lower layers in BERT than semantic information,
because syntactic information is more ``shallow.''
\citet{daiKnowledgeNeuronsPretrained2022} appear to agree with this assertion
in claiming that factual information is located in the upper layers.
\citet{mengLocatingEditingFactual2022}, however, claim
that factual information is located in the middle.  This contradiction may
support \cites{niuDoesBERTRediscover2022} assertion that
layers are not the best explanatory device of the distribution of these types of
information in LMs.
We explore here the possibility that no
dividing line exists at all between the mechanisms through which a language
model processes information related to these two types of competence.

\section{Localising Syntactic Phenomena in Language Models}
\label{sec:kn}

We put the KN thesis to the test under the KN-edit framework by asking
three questions:
(1) can we localise linguistic phenomena using the same KN-edit method;
(2) how do the levels of localisation compare to each other; and
(3) are these localisations strong enough to support the KN thesis?%
\footnote{Due to page restrictions, we only present the results
of the {determiner\_noun\_agreement\_2} (DNA.2) paradigm on BERT in the main
content. See Appendix \ref{app:kns_for_linguistic_phenomena} for the result of
other BLiMP paradigms and LMs.}

\subsection{Methods: Searching for KNs of Syntactic Phenomena}

For each prompt, we calculate an integral-of-gradient attribution score
$\alpha_i^{(l)}$ for the $i$-th intermediate neuron on the $l$-th layer
($w_i^{(l)}$).  Then, for a syntactic phenomenon with the source-target
pair $(s,t,p)$, we find the neurons that have an attribution score greater or
equal to $\pi$=20\% of the maximum attribution score shared among at least
$\tau$\% of its prompts.  We start from $\tau$=70\% and adjust it by an
increment or decrement of 5\% until the number of neurons is within the range of
$[2, 5]$.

\paragraph{Neuron Attribution Score}

Given an input prompt $x$, we follow \citet{daiKnowledgeNeuronsPretrained2022}
and use the integral of gradients to calculate the neuron attribution score:
\begin{equation}
\alpha_i^{(l)} = \overline{w}_i^{(l)}\int_{\gamma=0}^1
  \frac{\partial P_x(\gamma\overline{w}_i^{(l)})}
  {\partial w_i^{(l)}}d\gamma, \;
P_x(\hat{w}_i^{(l)}) = p(y|x,w_i^{(l)}=\hat{w}_i^{(l)}),
\end{equation}
where $P_x(\hat{w}_i^{(l)})$ denotes the probability distribution of the token
$y$ when changing the neuron $w_i^{(l)}$'s value to $\hat{w}_i^{(l)}$, and
$\frac{\partial P_x(\alpha\overline{w}_i^{(l)})}{\partial w_i^{(l)}}$ denotes
the gradient of the model with respect to the activation $w_i^{(l)}$.
We will see a more salient gradient when the neuron inflicts a greater change on
the output probability.

\begin{wrapfigure}{r}{0.6\linewidth}
\vspace{-2em}
\centering
\begin{subfigure}[b]{0.55\linewidth}
\includegraphics[width=\linewidth]{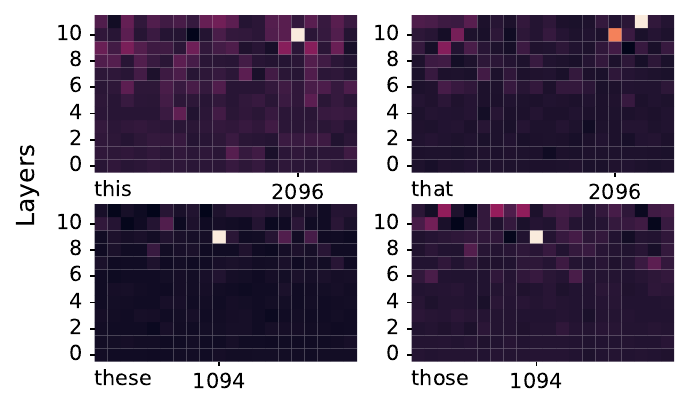}
\caption{Average KN attribution scores.}
\label{fig:avg_att_score}
\end{subfigure}\hfill
\begin{subfigure}[b]{0.42\linewidth}
\scriptsize\centering
\setlength\tabcolsep{2pt}
\raisebox{1.0\height}{
\begin{tabular}{c|c|c|c|c} \toprule
Neuron                   & this   & that   & these  & those  \\ \midrule
{\red $w^{(10)}_{2096}$} & 0.93   & 0.75   & 0      & 0      \\
{\blue $w^{(9)}_{1094}$} & 0      & 0      & 1.00   & 1.00   \\ \midrule
$w^{(9)}_{2339}$         & 0.33   & 0      & 0.32   & 0      \\
$w^{(11)}_{2686}$        & 0      & 0.81   & 0      & 0      \\
\ldots                   & \ldots & \ldots & \ldots & \ldots \\
\bottomrule
\end{tabular}}
\caption{KNs for Det-N pairs.}
\label{fig:neruon_dist}
\end{subfigure}
\vspace{-0.4em}
\caption{Localising grammatical number to KNs. The singular determiners 
share a common KN {\small($w_{2096}^{(10)}$)},
and the plural determiners share a different common KN {\small($w_{1094}^{
(9)}$)}.
}
\vspace{-1em}
\end{wrapfigure}

\paragraph{Measuring the Level of Localisation}
\label{par:measuring_the_level_of_localisation}

We use three metrics to measure the level of localisation:
(1) the number of identified neurons ($|\text{KN}|$) using the initial threshold
setting ($\tau$=70\%), 
(2) the final threshold $\tau$ to obtain 2-5 KNs,
and, (3) a similarity score among all the token attribution patterns.

Both of \cites{daiKnowledgeNeuronsPretrained2022} measures ($|\text{KN}|$ and $\tau$)
depend on adjusting the two threshold hyperparameters, $\pi$ and $\tau$.
Here, we propose a non-parametric measure using a generalised $n$-sample
similarity measure ($R_1^2$) that measures the correlation of all the
attribution patterns:
\begin{equation}
\begin{gathered}
  Y = [y_1 \ldots y_n], \; y_i = \frac{s_i}{\|s_i\|},
  Y = USV^\top = \sum_{k=1}^n \sigma_k u_k v_k^\top, \;
  R^2 = \frac{\sigma_1^2 - 1}{n - 1}. \\
\end{gathered}
\end{equation}
We first normalise and concatenate each attribution pattern $s_i$ for each prompt
$x_i$ in the dataset into
$Y$.  Then, we can calculate the
similarity/correlation among all $n$ patterns by conducting a singular value
decomposition (SVD) and using the square of the first singular value $\sigma_1^2$.
We then normalise 
this measure to the range $[0, 1]$ so that the similarity
between
$n$ parallel vectors will be $R_1^2=1$, and $n$ orthogonal vectors will get
$R_1^2=0$.

\subsection{Results \& Findings}

\paragraph{Finding 1: We can localise the grammatical number of determiners to
just two neurons, just like factual information.}
The BLiMP paradigm determiner\_noun\_agreement\_2 (DNA.2) contains 1000 sentence
pairs with exactly one demonstrative determiner ({\it this, that, these, those})
agreeing with an adjacent noun, e.g., {\it Carl cures 
\underline{those}/*\underline{that} \textit{\textbf{horses}}}.  The determiner 
{\it those} is $t$, {\it that} is $t^*$ and the noun {\it horses} is $s$.  A
noun may appear in multiple sentence pairs.  Among the paradigm's 1000 sentence
pairs, we identified 283 unique Det-N pairs $(s,t,t^*,r)$.

\textbf{Attribution Score Patterns}
The attribution score of neurons shows a highly consistent pattern that can be
interpreted linguistically.  We calculated the average
attribution scores of all the prompts that contains each one of the determiners.
Figure \ref{fig:avg_att_score} shows a selection of the average attribution
scores.  The colour block in the $i$th column and $j$th row shows the
attribution score {\small $\alpha_i^{(j)}$}.  As we can see, a common neuron
{\small($w_{2096}^{(10)}$)} has a high average attribution score for both of the
singular determiners {\it this} and
{\it that}, and another common neuron
{\small($w_{1094}^{(9)}$)} lights up for the plural determiners {\it these} and
{\it those}.\footnote{We use \cites{mengLocatingEditingFactual2022} neuron
numbering system.  Both layer and neuron indices start with 0.}

This pattern is not only shown in aggregate.  For each Det-N pair, we use the
1000 sentences in the paradigm as templates to create the prompts needed for a
KN search.  For each sentence, we replace the sentence's determiner and noun
with the Det-N's determiner and noun. We then obtain 1000 sentences with
different contexts but the same determiners and nouns.  Then, we run a KN search
on these 1000 sentences.
When we look into each individual Det-N pair, the two neurons are identified as
KNs in the vast majority of the pairs.  As shown in Figure 
\ref{fig:neruon_dist}, $w_{2096}^{(10)}$ appeared
in 93\% of the pairs with {\it this} and 75\% of the pairs with {\it that}. 
The plural neuron appeared in 100\% of pairs with {\it these} or {\it those}. 
More importantly, these neurons were not identified as KNs in pairs with the
opposite grammatical numbers.  Figure \ref{fig:neruon_dist} shows an excerpt of
the results (full results in Appendix \ref{app:the_kn_search}). 

\begin{figure}
\includegraphics[width=0.5\linewidth]
{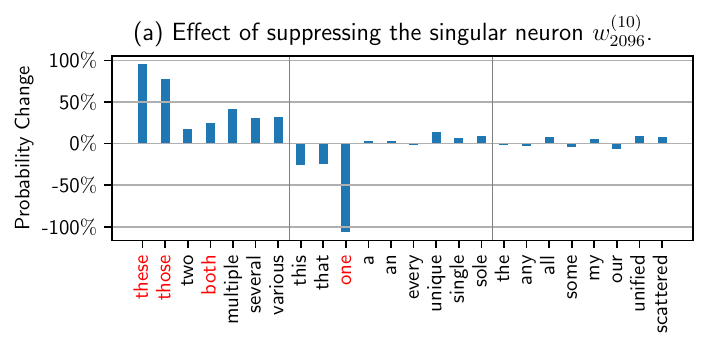} \hfill
\includegraphics[width=0.5\linewidth]
{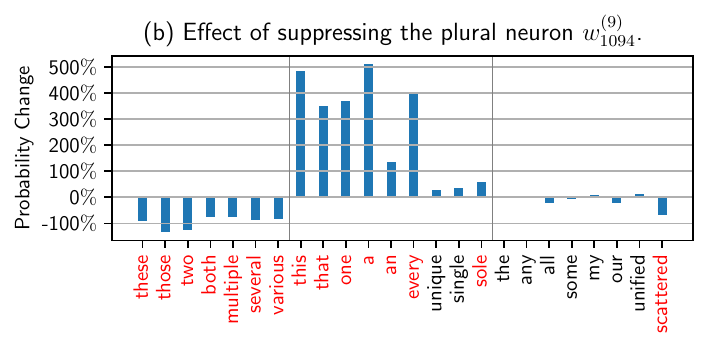}
\vspace{-2em}
\caption{Suppressing the number neuron's (singular: $w^{(10)}_{2096}$; plural:
$w^{(9)}_{1094}$) effect across
number-expressing prenominal modifiers.  Significant ($p<0.05$) changes are
highlighted in {\red red}.  The three sections in the plots are, from left to
right, plural, singular and neutral modifiers.}
\label{fig:erase_effect}
\vspace{-4em}
\end{figure}

\paragraph{Effects of Suppressing the ``Number Neuron''}
Do these two neurons correspond to grammatical number?
We suppress each neuron (setting activation to 0) and compute the pre- and
post-edit model's output probability of various number-expressing prenominal
modifiers across all prompts with singular/plural nouns.  Appendix 
\ref{sub:prenominal_modifiers} explains the prenominal modifier selection
process. Figure \ref{fig:erase_effect} shows the average effect
of suppressing the identified KNs
($\frac{p(\text{post-edit})-p(\text{pre-edit})}
{\min (p(\text{post-edit}), p(\text{pre-edit}))}$).

\begin{wrapfigure}{r}{0.25\linewidth}
\vspace{-1.3em}
\includegraphics[width=\linewidth]{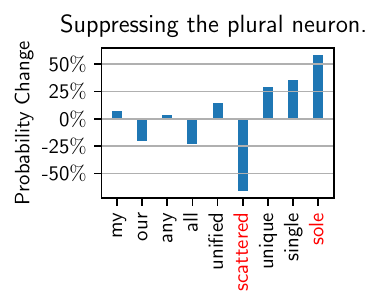}
\vspace{-2em}
\caption{The localisation of plurality appeals to word co-occurrence
frequencies cues.}
\label{fig:zoom-in}
\vspace{-1em}
\end{wrapfigure}
The result of suppressing the plural neuron is pronounced (Figure 
\ref{fig:erase_effect}b).
This intervention leads to a
significant reduction in
probability across all plural modifiers, a notable increase for the majority of
singular modifiers, but a limited impact for modifiers that do not express number agreement.  Therefore, erasing the activation of the plural neuron
causes a decrease in the expression of determiner-noun agreement for plural
modifiers. Although this KN search is solely based on these four demonstrative
determiners, we observed that it generalizes to other determiners ({\it one,
a, an, every;}
{\it two, both; multiple, several, various}) and even adjectives 
({\it single, unique, sole}).
This effect is statistically significant.  By treating the pre- and post-edit
probabilities as two separate groups, a
\cites{studentProbableErrorMean1908} $t$-test reveals significance
when the modifiers are highlighted in
{\red red} in Figure \ref{fig:erase_effect}.
The null hypothesis is that the pre- and post-edit
probabilities are sampled from the same distribution, i.e., the intervention has
no effect. Thus, the neuron $w^{(9)}_{1094}$ can be interpreted through the lens
of a linguistic phenomenon, viz.\ determiner-noun agreement.

Note, however, that the word {\it
scattered} also sees a significant probability decrease when suppressing the
plural neuron.  {\it Scattered} does not specify for plural number; phrases such as ``scattered rioting'' are syntactically and
semantically well-formed.  But it is used more often
with plural nouns because of its meaning. This frequency effect is not limited to {\it scattered}.  Other
words such as {\it any,
all, unified}, and the three adjectives {\it unique, single} and {\it
sole} exhibit a similar bias.  As shown
in Figure \ref{fig:zoom-in},
we see probability changes, although less substantial, alongside
those modifiers that strictly specify for grammatical number.  This
is a semantic number co-occurrence bias.

The suppression effect of the singular neuron is similar but less
pronounced.  Overall, we see the opposite effect across all prenominal
modifiers, with the ``singular'' adjectives ({\it unique, single, sole}) being
the only exceptions.  This is, however, unsurprising.  Unlike the plural
neuron, the singular neuron did not appear in all of the Det-N pairs.  We
suspect that an LM can identify the plural property more easily when its
wordpiece-based tokeniser exposes many plural suffixes.
\begin{figure}
\begin{subfigure}[b]{0.53\linewidth}
\centering\scriptsize\setlength\tabcolsep{3pt}
    \begin{tabular}{l|c|c|c} \toprule
    BLiMP Paradigm         & $|\text{KN}|$ & $\tau$ & $R_1^2$ \\ \midrule
    det\_n\_agr.\_1        & 3.94          & 0.71   & 0.56    \\
    det\_n\_agr.\_2        & 1.86          & 0.62   & 0.56    \\
    dna.\_irr.\_1          & 5.53          & 0.73   & 0.64    \\
    dna.\_irr.\_2          & 2.45          & 0.67   & 0.55    \\
    dna.\_w.\_adj\_1       & 8.88          & 0.78   & 0.67    \\
    dna.\_w.\_adj\_2       & 2.26          & 0.67   & 0.57    \\
    % dna.\_w.\_adj\_irr.\_1 & 9.79          & 0.78   & 0.67    \\
    % dna.\_w.\_adj\_irr.\_2 & 2.60          & 0.69   & 0.58    \\
    \bottomrule
    \end{tabular} \hfill
    \begin{tabular}{l|c|c|c} \toprule
    Rels. & $|\text{KN}|$ & $\tau$ & $R_1^2$ \\ \midrule
    P101  & 0.167         & 0.515  & 0.399   \\
    P103  & 0.204         & 0.662  & 0.399   \\
    P106  & 1.292         & 0.607  & 0.365   \\
    P108  & 1.493         & 0.663  & 0.473   \\
    P1303 & 10.462        & 0.814  & 0.684   \\
    P140  & 2.008         & 0.689  & 0.263   \\
    % P1412 & 2.196         & 0.687  & 0.612   \\
    % P19   & 2.597         & 0.693  & 0.481   \\
    \bottomrule
    \end{tabular}
\caption{Levels of localisation measures.}
\label{fig:level_localisation}
\end{subfigure}
\hfill
\begin{subfigure}[b]{0.46\linewidth}
\centering
\includegraphics[width=0.48\linewidth]{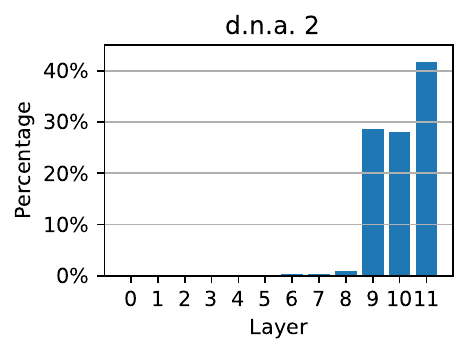}
\includegraphics[width=0.48\linewidth]{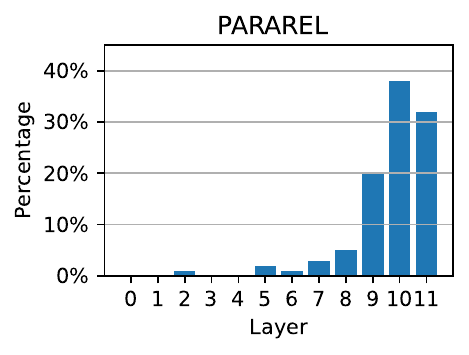}
\vspace{-0.5em}
\caption{Layer distribution of identified KNs.  Both BLiMP and {\sc ParaRel} occupy
the topmost layers.}
\label{fig:layer_dist}
\end{subfigure}
\caption{The localisation of certain syntactic phenomena (BLiMP) is comparable to facts
({\sc ParaRel}).  We see comparable localisation metrics and the identified KNs
occupy the same layers.}
\end{figure}

\begin{figure}
\vspace{-0.7em}
\begin{subfigure}[b]{0.3\linewidth}
    \centering\scriptsize
    \setlength\tabcolsep{2pt}
    \includegraphics[width=\linewidth]{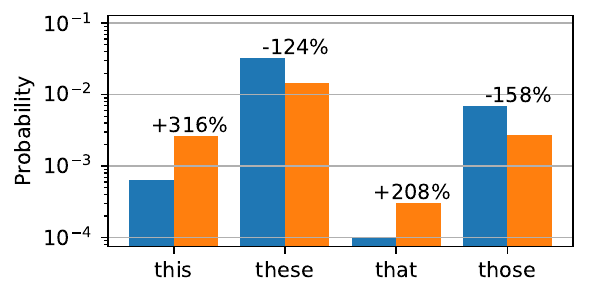}
\vspace{-2em}
\caption{The exact effect to output probability of editing the KNs.
\raisebox{0.5\height}{\colorbox{bblue}{\makebox(0.2ex,0.2ex){}}}: pre-edit.
\raisebox{0.5\height}{\colorbox{orange}{\makebox(0.2ex,0.2ex){}}}: post-edit.}
\end{subfigure}\hfill
\begin{subfigure}[b]{0.36\linewidth}
    \centering\scriptsize\setlength\tabcolsep{3pt}
    % \vspace{1ex}
    \begin{tabular}{l|c|c|c} \toprule
    Paradigm                & Pre-edit & Post-edit & $\Delta$ \\ \midrule
    det\_n\_agr.\_2         & 100\%    & 94.8\%    & -5.2\%   \\
    dna.\_irr.\_2           & 99.5\%   & 96.9\%    & -2.6\%   \\
    dna.\_w.\_adj.\_2       & 97.1\%   & 94.4\%    & -2.7\%   \\
    dna.\_w.\_adj.\_irr.\_2 & 97.4\%   & 95.4\%    & -2.0\%   \\
    \bottomrule
    \end{tabular}
\caption{These modifications of determiner-noun KNs are usually not enough to
overturn the categorical prediction.}
\end{subfigure}\hfill
\begin{subfigure}[b]{0.3\linewidth}
    \centering\scriptsize
    \begin{tabular}{l|c|c} \toprule
    Data                         & Model & Reliability \\ \midrule
    \multirow{2}{*}{ZsRE}        & T5-XL & 22.51       \\
                                 & GPT-J & 11.34       \\ \midrule
    \multirow{2}{*}{CounterFact} & T5-XL & 47.86       \\
                                 & GPT-J & 1.66        \\
    \bottomrule
    \end{tabular}
\caption{KN edit has low reliability for facts \citep{yaoEditingLargeLanguage2023}.}
\label{fig:kn_is_bad}
\end{subfigure}

\caption{Editing the KNs is not enough to overturn the categorical
predictions.  The major limitation of KN edit is its low reliability.  These
reliability scores cannot support the KN thesis.}
\label{fig:ee_exact_numbers}
\vspace{-4em}
\end{figure}

\paragraph{Finding 2: KNs obtained using linguistic tasks and factual tasks
share similar characteristics of localisation.}
Figure \ref{fig:level_localisation} shows the level of localisation of various
BLiMP determiner-noun
agreement paradigms and selected {\sc ParaRel} relations.  The
localisation metrics of both BLiMP paradigms and {\sc ParaRel} relations fall within the
same range.  See Appendix \ref{sub:level_of_localisation} for the full list.

Furthermore, Figure \ref{fig:layer_dist} shows no bifurcation of
layers within which linguistic and factual KNs locate
(see Appendix \ref{app:layer_dist_more}).  All of the neurons are distributed in
the topmost layers.  The determiner-noun agreement pattern is purely syntactic. 
This is a refutation of \citet{jawaharWhatDoesBERT2019} and
\cites{tenneyBERTRediscoversClassical2019} view that syntax is localised to
more shallow layers than
semantics.  Our results confirm \cites{niuDoesBERTRediscover2022}
assertion that the location of syntactic and semantic (and, additionally,
factual) information is not distinguished by
layer in the LM.  In fact, our results may suggest that these types of
information are most fruitfully thought of as being handled by the same 
functional mechanism.

\paragraph{Finding 3: Despite the high level of localisation in the underlying
probability drift, the effect of editing the KNs is not enough to
overturn the
categorical predictions made by the language model.}
Although we see a high level of localisation in the relative probability change
between $t$ and $t,^*$ we find that this change is often
not enough to overturn the final prediction.  As shown in Figure
\ref{fig:ee_exact_numbers}, we only see at most 5.2\% of the BLiMP
results being overturned.  This low reliability issue is not limited to
syntactic phenomena.  In Figure \ref{fig:kn_is_bad}, we list 
\cites{yaoEditingLargeLanguage2023} evaluation of KN edit on two other corpora:
ZsRE \citep{levyZeroShotRelationExtraction2017} and CounterFact 
\citep{mengLocatingEditingFactual2022}.  The reliability of the KN algorithm
ranges from 1.66\% to 47.86\% --- not enough to support the KN thesis.

\paragraph{Discussion}
\label{par:discussion}
Just as with facts, syntactic phenomena localise to neurons.
Modifying merely two neurons working in tandem can significantly change the
expression of determiner-noun number.  This is not the only type of
localisable syntactic phenomenon (see Appendix
\ref{app:kns_for_linguistic_phenomena}), and together they constitute a significant extension of
\cites{finlaysonCausalAnalysisSyntactic2021} findings --- syntactic
phenomena can be localised to the individual neuron level.  
Furthermore, these phenomena share with
factual information the extent of their localisation, and the layers in which 
the KNs typically occur.

But do the patterns identified for these neurons constitute
``knowledge?''  KN edit's low reliability score
and its appeal to shallow cues both suggest otherwise.  If we follow the KN thesis and
interpret a post-edit probability change as an indication of the quantity of
knowledge stored, then we cannot draw the conclusion that knowledge is stored there.
The identified neurons are spots with a high information concentration, but the
final decision still lies with the rest of the model.

Interestingly, the patterns that we identified resemble linguistic
categories, but they deviate from rules of grammatical well-formedness.  In determiner-noun agreement, KN edit also affects pre-modifiers that do not 
specify for number, alongside plural-specifying determiners such as 
{\it multiple}, {\it several} and {\it various}.
Phrases such as {\it sole breadwinners} and {\it
scattered rioting} are less frequent but by no means unheard of.
This suggests that the patterns reflected within the MLP neurons can only
be completely accounted for by appealing
to superficial cues such as word co-occurrence frequency.

\section{Causal Tracing and Rank-One Model Editing}
\label{sec:rome}

\begin{wrapfigure}{r}{0.65\linewidth}
\vspace{-1.5em}
\centering
\begin{subfigure}[b]{\linewidth}
\includegraphics[width=\linewidth]{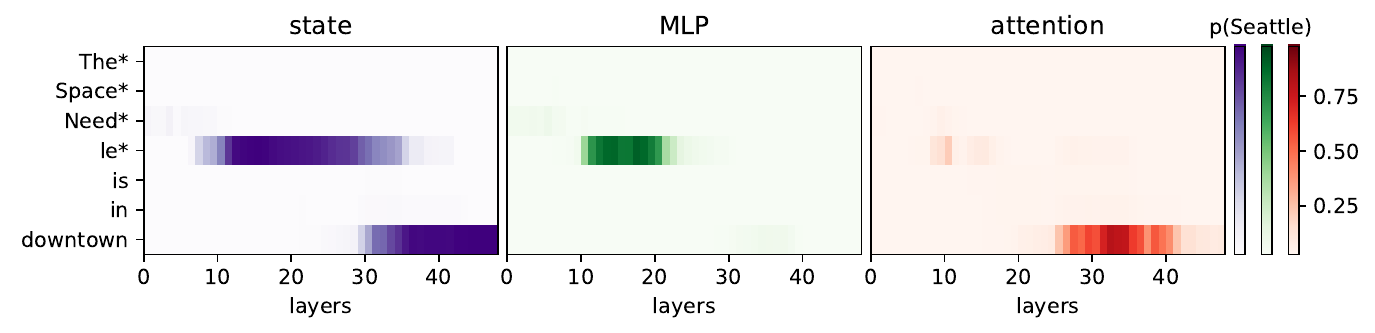}
\vspace{-1.5em}
\caption{Factual information.}
\label{fig:ct_fact}
\end{subfigure}

\begin{subfigure}[b]{\linewidth}
\includegraphics[width=\linewidth]{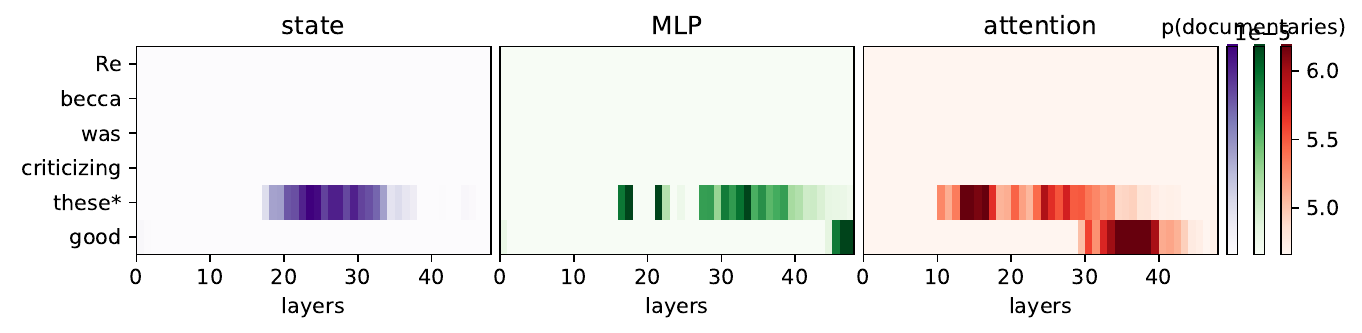}
\vspace{-1.5em}
\caption{Determiner-noun agreement.}
\end{subfigure}

\begin{subfigure}[b]{\linewidth}
\includegraphics[width=\linewidth]{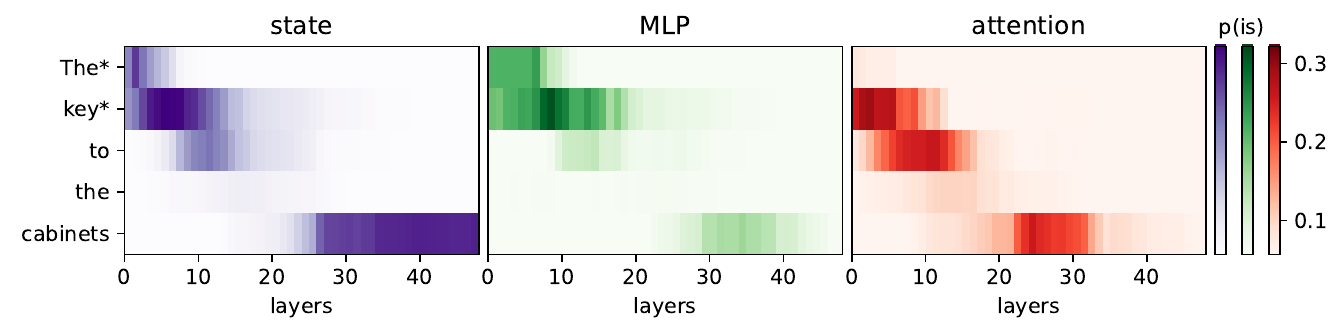}
\vspace{-1.5em}
\caption{Subject-verb agreement.}
\end{subfigure}
\vspace{-2em}
\caption{Causal tracing result.}
\vspace{-1em}
\label{fig:ct}
\end{wrapfigure}
% \vspace{-4em}

In this section, we reassess \cites{mengLocatingEditingFactual2022} similar but
more intricate implementation of KN edit.  They proposed that information
is expressed at two locations: facts are recalled in mid-layer MLP weights, and
copied to the final output by attention modules.  They derived this
thesis based on causal mediation.  The causal traces in Figure
\ref{fig:ct_fact} are computed as follows.  First, the source tokens are corrupted
by adding random noise $\epsilon$ and the model generates an
incorrect result.  Then, they restore an intermediate hidden state to its
correct value for all the tokens at all layers, and determine whether this restoration
can fix the corruption.  They discover a division of labour between the MLP and attention.
This division, however, is not stable.  In Figure 
\ref{fig:ct}bc we reproduce this effect on syntactic phenomena.  The
distinction between the early and late site is no longer discernible.  This is,
in fact, not a distinction between facts and syntactic patterns.  Many factual
causal traces also do not show this distinction.%
\footnote{In
Appendix \ref{app:causal_tracing}, we show the average indirect causal effects;
our observation thus holds under aggregation.}  

Previous evaluation of the ROME model-editing method was limited to
simple paraphrasing \citep{yaoEditingLargeLanguage2023}.  We observe that ROME
does not generalise well in respect of either of our new criteria, bijective
symmetry or synonymous invariance (Figure \ref{fig:rome_edit}ab). 
This issue persists when we evaluate ROME quantitatively.  We assembled two new
datasets using {\sc ParaRel} relations to evaluate our two new criteria
(see Appendix \ref{sec:extra_evaluation} for details). We use
the two bijective relations R1376 and P36 to construct a bijective symmetry evaluation
dataset.  Then, for synonymous invariance, we rewrite the field-of-work targets in P101 into
occupation names.  For instance, if we change Anaxagoras's field of work from 
{\it philosophy} to {\it linguistics}, we also want the model to answer
``Anaxagoras is a \underline{linguist}'' when given the prompt.
Table \ref{tab:extra_eval} shows the result of our evaluation on these newly
assembled datasets.  Although ROME obtains higher reliability scores than KN edit
in both GPT-2 XL and LLaMA-2 7B, the symmetry and synonymy results are both much lower.
We also observe that ROME edit can only edit the exact association between the
tokens in $(s,t,r)$.  As demonstrated in Figure \ref{fig:rome_edit}c, editing
the verb corresponding to {\it the authors} from {\it are} to {\it is}
only affects the subject {\it the authors}, and not other subjects such as 
{\it the pilots}.
These look more like at-times brittle
\begin{wraptable}{r}{0.37\linewidth}
\vspace{-0.2em}
\caption{Results obtained under our new criteria suggest model editing methods
are not robust.}
\vspace{-0.7em}
\scriptsize\centering\setlength\tabcolsep{2pt}
\begin{tabular}{llclc} \toprule
Model    & Data  & Reliability & \multicolumn{2}{c}{Measure} \\ \midrule
         & P101  & 99.82\% & Synonym  & 52.35\%  \\
GPT-2 XL & P1376 & 96.37\% & Symmetry & 23.71\% \\
         & P36   & 99.79\% & Symmetry & 25.17\% \\ \midrule
         & P101  & 100\% & Synonym  & 58.36\%  \\
LLaMA-2  & P1376 & 100\%   & Symmetry & 33.40\% \\
         & P36   & 100\%   & Symmetry & 33.64\% \\
\bottomrule
\end{tabular}
\label{tab:extra_eval}
\vspace{-5em}
\end{wraptable} 
patterns of token expression than factual knowledge.

\begin{figure}
\centering\tiny
\begin{tabular}{p{4cm}} \toprule
(a) {\bf GPT-2 XL}: {\it The capital of \underline{Canada} is} {\green Ottawa} \\
{\bf ROME Edit}: Ottawa $\rightarrow$ Rome \\ \midrule

\original: {\it The capital of \underline{Canada} is} {\green Ottawa} ... \\
\patched: {\it The capital of \underline{Canada} is} {\red Rome}. \\ \midrule

\original: {\it \underline{Ottawa} is the capital of} {\green Canada}. \\
\patched: {\it \underline{Ottawa} is the capital of} {\green Canada}'s
federalist system of government. \\ \midrule

\original: {\it \underline{Rome} is the capital of} {\green Italy}, ... \\
\patched: {\it \underline{Rome} is the capital of} {\green Italy}, ... \\ 

\bottomrule
\end{tabular}
\hfill
\begin{tabular}{p{4.1cm}} \toprule
(b) {\bf GPT-2 XL}: {\it To treat my \underline{toothache}, I should see a} 
{\green dentist} \\
{\bf ROME Edit}: dentist $\rightarrow$ lawyer \\ \midrule

\original: {\it To treat my \underline{toothache}, I should see a}
{\green dentist}, ... \\
\patched: {\it To treat my \underline{toothache}, I should see a} {\red lawyer}.\\
\midrule

\original: {\it To treat my \underline{tooth pain}, I should see a}
{\green dentist}.\\
\patched:  {\it To treat my \underline{tooth pain}, I should see a}
{\green dentist}. \\ \midrule

\original: {\it To treat my \underline{odontalgia}, I should see a}
{\green dentist}.\\
\patched:  {\it To treat my \underline{odontalgia}, I should see a}
{\green dentist}. \\
\bottomrule
\end{tabular}
\hfill
\begin{tabular}{p{4.5cm}} \toprule
(c) {\bf GPT-2 XL}: {\it \underline{The authors} near the taxi drivers} {\green
are} \\
{\bf ROME Edit}: are $\rightarrow$ is \\ \midrule

\original: {\it \underline{The authors} near the taxi drivers}
\underline{\green are} ... \\
\patched: {\it \underline{The authors} near the taxi drivers}
\underline{\red is} ... \\ \midrule

\original: {\it \underline{The authors} near the dancers} in their paper
\underline{\green are} ... \\
\patched:  {\it \underline{The authors} near the dancers} {\red is} ... \\
\midrule

\original: {\it \underline{The pilots} near the taxi drivers}
{\green were} ... \\
\patched: {\it \underline{The pilots} near the taxi drivers}' cabins
{\green are} ... \\ \midrule

\original: {\it \underline{The pilots} near the dancers} {\green are} ... \\
\patched: {\it \underline{The pilots} near the dancers} {\green are} ... \\
\bottomrule
\end{tabular}

\caption{Comparison of generated text. The prompts are {\it italicized}, \underline{source tokens} ($s$) are underlined,
ungrammatical or counter-factual responses are highlighted in {\red red}, and
unchanged correct responses in {\green green}.  
\original~shows the original GPT-2 XL's generation, and \patched~shows the
edited model's response.}
\label{fig:rome_edit}
\vspace{-6em}
\end{figure}

\section{Discussion \& Conclusion}
\label{sec:discussion}

We find that several syntactic agreement phenomena can be localised to a small number of
MLP neurons.  This localisation has similar characteristics to the localisation of factual
information, suggesting that recent transformer-based language models'
impressive abilities with respect to various linguistic phenomena and the recall of
facts from their training corpora may follow the same underlying mechanism. 
The localisation of the two types of information also faces the same
challenges, however, which militate against the soundness of the KN thesis.  Specifically,
the effect of editing the identified neurons is not strong enough to overturn the
final prediction, and the scope of the phenomena appears to be limited to shallow cues such
as token co-occurrence statistics.

Returning to \cites{gevaTransformerFeedForwardLayers2021}
original findings, the MLP neurons store patterns that are
interpretable through a linguistic lens, but they do not store
knowledge, either linguistic or factual.  
\cites{mengLocatingEditingFactual2022} causal tracing results, although still
an oversimplification, suggest that there are different phases in
different layers in the entire process of token expression.  But their ROME 
model-editing method did not avail itself of this important finding.  The method is
still MLP-based.  To achieve a better understanding of this
expression process and achieve real model editing, we must examine the entire
decision-making circuit 
\citep{wangInterpretabilityWildCircuit2022,wuInterpretabilityScaleIdentifying2023,conmyAutomatedCircuitDiscovery2023,murtyCharacterizingIntrinsicCompositionality2023}.  Manipulating only the MLP weights is not enough. The circuit mode of interpretation is still at a very early state of development, however. 
Current circuit identification methods are \textit{ad hoc}, furthermore, and have only been applied to a small set of tasks. In future work, we will try to formalize the circuit interpretation framework and apply it to more tasks and phenomena.

Our reassessment of causal
traces agrees with \cites{haseDoesLocalizationInform2023}, but we
take exception to their claim that ``better mechanistic understanding \ldots may not
always translate to insights about how to best change their behavior.''
It is well-established that we can interpret the computational mechanism of LMs
through the lens of formal linguistics \citep{clarkWhatDoesBERT2019}.
Both of our findings reveal limitations in current LM
interpretation work and suggest that an even more comprehensive, but still
mechanistic interpretation of
transformers will lead to insights for better control of model behaviour
when not limited to the MLP modules, and when patterns of token expression
are dealt with unencumbered by misbegotten metaphors about knowledge and human
reasoning.

\paragraph{Contributions}
Our work provides a thorough examination of the KN thesis and finds that the
thesis is, at best, an oversimplification.  We (1) extend KN-based
analysis to well-defined syntactic tasks, (2) propose two new criteria
for evaluating the effectiveness of model editing, and (3) introduce a generalised $n$-sample similarity measure of the level of localisation.

\subsubsection*{Acknowledgments}
We thank Lei Yu (University of Toronto) for a great deal of insightful discussion.
We also want to thank the anonymous reviewers for providing informative
comments and suggestions.

\bibliography{iclr2024_conference}
\bibliographystyle{iclr2024_conference}

\newpage
\appendix

\section{Experimental Setup}
\label{sec:experimental_setup}

\subsection{Language Models}
\label{sub:language_models}

We experiment on BERT \citep{devlinBERTPretrainingDeep2019}, GPT-2 
\citep{radfordLanguageModelsAre2019}, and LLaMA-2 
\citep{touvronLLaMAOpenEfficient2023}.  We use the {\tt bert-base-cased} version
of BERT, the base version GPT-2 and 7B parameter version of LLaMA-2 in Section
\ref{sec:kn}.  We added GPT-2 XL in Section \ref{sec:rome} as it is also used by
\citep{mengLocatingEditingFactual2022}.  We use the huggingface package 
\citep{wolfHuggingFaceTransformersStateoftheart2020} for the implementation.

We choose BERT and GPT-2 as they are the most widely studied and applied
language models.  We choose LLaMA-2 as a representative of the recent large
language models.  All three models are transformer-based.  BERT is a masked
language model (MLM), and GPT-2 and LLaMA are decoder-only language
models.  MLM is a type of bidirectional language model.  It can process context
in both the forward and backward direction and the order between source and
target is not required.  However, decoder-only language models only process
context from left to right and this means that the target must locate at the end
of the prompt.  Therefore, some linguistic patterns are not suitable for
decoder-only LMs.  We discarded these patterns for GPT-2 and LLaMA.

For the sake of consistency with \citet{mengLocatingEditingFactual2022}, we use
GPT-XL for causal tracing and the evaluation of ROME.  We also evaluate LLaMA-2-7B
on ROME as a representative of recent LLMs.  \citet{yaoEditingLargeLanguage2023}
provided a recipe of applying ROME on LLaMA-2-7b,%
\footnote{\href{https://github.com/zjunlp/EasyEdit/blob/main/hparams/ROME/llama-7b.yaml}{https://github.com/zjunlp/EasyEdit/blob/main/hparams/ROME/llama-7b.yaml}}
we follow their instructions and hyperparameters to conduct our evaluation.

\subsection{Corpora}

\paragraph{BLiMP}
\label{par:blimp}

\begin{table}
\centering\scriptsize
\caption{BLiMP phenomena and paradigms.}
\begin{tabular}{l|l|l} \toprule
Phenomenon                   & Paradigms                                            & Example                                                                             \\ \midrule
Anaphor                      & anaphor\_gender\_agreement                           & Katherine can't help \underline{herself}/*\underline{himself}.                      \\
Agreement                    & anaphor\_number\_agreement                           & Many teenagers were helping \underline{themselves}/*\underline{herself}.            \\ \midrule
                             & determiner\_noun\_agreement\_1                       & Craig explored that \underline{grocery store}/*\underline{grocery stores}.          \\
                             & determiner\_noun\_agreement\_2                       & Carl cures \underline{those}/*\underline{that} horses.                              \\
\multirow{2}{*}{Determiner-} & determiner\_noun\_agreement\_irregular\_1            & Phillip was lifting \underline{this mouse}/*\underline{this mice}.                  \\
\multirow{2}{*}{Noun}        & determiner\_noun\_agreement\_irregular\_2            & Those ladies walk through \underline{those}/*\underline{that} oases.                \\
\multirow{2}{*}{Agreement}   & determiner\_noun\_agreement\_with\_adj\_1            & Tracy praises those lucky \underline{guys}/*\underline{guy}.                        \\
                             & determiner\_noun\_agreement\_with\_adj\_2            & Some actors buy \underline{these}/*\underline{this} gray books.                     \\
                             & determiner\_noun\_agreement\_with\_adj\_irregular\_1 & This person shouldn't criticize this upset \underline{child}/*\underline{children}. \\
                             & determiner\_noun\_agreement\_with\_adj\_irregular\_2 & That adult has brought \underline{that}/*\underline{those} purple octopus.          \\ \midrule
                             & distractor\_agreement\_relational\_noun              & A sketch of lights \underline{doesn't}/*\underline{don't} appear.                   \\
\multirow{2}{*}{Subject-}    & distractor\_agreement\_relative\_clause              & Boys that aren't disturbing Natalie \underline{suffer}/*\underline{suffers}.        \\
\multirow{2}{*}{Verb}        & irregular\_plural\_subject\_verb\_agreement\_1       & This goose \underline{isn't}/*\underline{weren't} bothering Edward.                 \\
\multirow{2}{*}{Agreement}   & irregular\_plural\_subject\_verb\_agreement\_2       & The \underline{woman}/*\underline{women} cleans every public park.                  \\
                             & regular\_plural\_subject\_verb\_agreement\_1         & Jeffrey \underline{hasn't}/*\underline{haven't} criticized Donald.                  \\
                             & regular\_plural\_subject\_verb\_agreement\_2         & The \underline{dress}/*\underline{dresses} crumples.                                \\
\bottomrule
\end{tabular}
\label{tab:blimp_overview}
\end{table}

We use the linguistic phenomena collected in BLiMP 
\citep{warstadtBLiMPBenchmarkLinguistic2020} for our analysis.
The BLiMP corpus contains minimal pairs for 12 grammar phenomena.  Some of the
phenomena are not suitable for our experiments and are therefore discarded.
The remaining phenomena and paradigms are shown in Table 
\ref{tab:blimp_overview}.

\paragraph{ParaRel}
\label{par:pararel}

The corpus {\sc ParaRel} \citep{elazarMeasuringImprovingConsistency2021}
contains facts formulated as a fill-in-the-blank cloze task and it is curated by
experts.  It contains 38 relation types and Table \ref{tab:pararel}
provides an overview of an overview of the {\sc ParaRel} corpus, and 27,738
relational facts in total.  We obtain prompts from {\sc ParaRel} following
\cites{daiKnowledgeNeuronsPretrained2022} instructions.  For each {\sc ParaRel}
relations, \citet{daiKnowledgeNeuronsPretrained2022} created multiple prompt
templates.  On average, they created 8.63 different prompt templates for each of
the relations.  In total, the {\sc ParaRel} corpus contains 253,448 prompts.

There are two bijective (1-1) relations: P1376 (capital of) and P36 (capital). 
We use those two relations for our bijection reversal relation evaluation.

\begin{table}
\small\centering
\caption{An overview of {\sc ParaRel} relations.  There are two bijective
relations: P1376 (capital of) and P36 (capital).  Both relations are
underlined.}
\begin{tabular}{l|l|c} \toprule
ID                & Relation                                         & Relation type   \\ \midrule
P1001             & applies to jurisdiction                          & N-M             \\
P101              & field of work                                    & N-M             \\
P103              & native language                                  & N-1             \\
P106              & occupation                                       & N-M             \\
P108              & employer                                         & N-M             \\
P127              & owned by                                         & N-1             \\
P1303             & instrument                                       & N-M             \\
P131              & located in the administrative territorial entity & N-1             \\
P136              & genre                                            & N-1             \\
\underline{P1376} & \underline{capital of}                           & \underline{1-1} \\
P138              & named after                                      & N-1             \\
P140              & religion                                         & N-1             \\
P1412             & languages spoken, written or signed              & N-M             \\
P159              & headquarters location                            & N-1             \\
P17               & country                                          & N-1             \\
P176              & manufacturer                                     & N-1             \\
P178              & developer                                        & N-M             \\
P19               & place of birth                                   & N-1             \\
P190              & twinned administrative body                      & N-M             \\
P20               & place of death                                   & N-1             \\
P264              & record label                                     & N-1             \\
P27               & country of citizenship                           & N-M             \\
P276              & location                                         & N-1             \\
P279              & subclass of                                      & N-1             \\
P30               & continent                                        & N-1             \\
\underline{P36}   & \underline{capital}                              & \underline{1-1} \\
P361              & part of                                          & N-1             \\
P364              & original language of film or TV show             & N-1             \\
P37               & official language                                & N-1             \\
P39               & position held                                    & N-M             \\
P407              & language of work or name                         & N-1             \\
P413              & position played on team / speciality             & N-1             \\
P449              & original network                                 & N-1             \\
P463              & member of                                        & N-M             \\
P47               & shares border with                               & N-M             \\
P495              & country of origin                                & N-1             \\
P530              & diplomatic relation                              & N-M             \\
P740              & location of formation                            & N-1             \\
P937              & work location                                    & N-M             \\
\bottomrule
\end{tabular}
\label{tab:pararel}
\end{table}

\paragraph{CounterFact and ZsRE}
\label{par:counterfact_and_zsre}

\citet{mengLocatingEditingFactual2022} processed {\sc ParaRel} relations
differently from \citet{daiKnowledgeNeuronsPretrained2022}.  In particular, they
did not create the new prompt templates for each relational facts.

The Zero-Shot Relation Extraction (zsRE) corpus used by 
\citet{mitchellFastModelEditing2022, decaoEditingFactualKnowledge2021} is
another popular corpus used to evaluate model editing methods.  The evaluation
slice contains 10,000 records.

Both corpora are used by \citet{mengLocatingEditingFactual2022} and
\citet{yaoEditingLargeLanguage2023}.  We did not use these two corpus in our
experiments, however, we are citing results conducted by 
\citet{mengLocatingEditingFactual2022} and \citet{yaoEditingLargeLanguage2023}
to avoid duplication.
Our analysis of factual information can be easily generalised to these two
corpora and we plan to expand our analysis to more corpora for future work.

\section{Determiner Grammatical Number KN Search}
\label{sec:determiner_grammatical_number_kn_search}

\subsection{Prenominal Modifiers} % (fold)
\label{sub:prenominal_modifiers}

We study the following prenominal modifiers:

\begin{itemize}
    \item Determiners:
    \begin{itemize}
        \item The demonstrative determiners used by BLiMP: {\it this, that,
        these, those};
        \item Plural determiners: {\it two, both, multiple, several, various};
        \item Singular determiners: {\it one, a, an, every};
        \item Determiners that do not express number agreement: {\it the, some,
        \underline{any, all, my, our}};
    \end{itemize}
    \item Adjectives: {\it single, unique, sole, \underline{scattered,
    unified}}.
\end{itemize}

The six modifiers that do not express grammatical number agreement:
{\it any, all, my, our} {\it scattered} and {\it unified} are underlined. 
However, these modifiers are used more often with numbered nouns because of
their meanings.

\subsection{The KN Search}
\label{app:the_kn_search}

Table \ref{tab:neurons_identified} shows the full list of neurons identified for the paradigm.
We identified two knowledge neurons.  The plural neuron $w^{9}_{1094}$ is
highlighted in {\blue blue} and the singular neuron $w^{10}_{2096}$ is highlighted in
{\red red}.

\begin{figure}
\includegraphics[width=0.5\linewidth]
{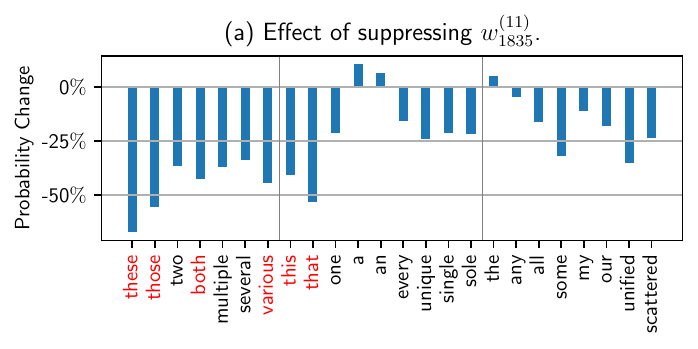} \hfill
\includegraphics[width=0.5\linewidth]
{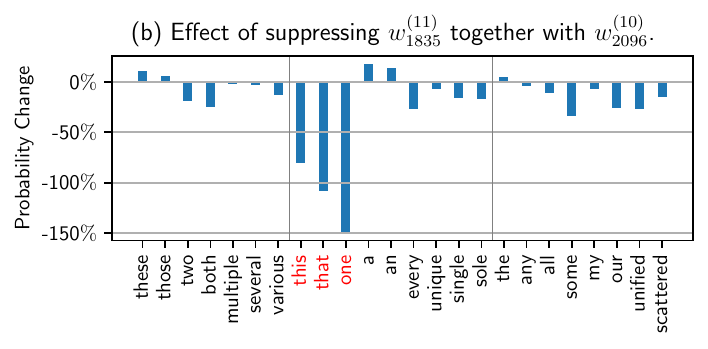}
\caption{Despite $w^{11}_{1835}$'s high occurrence in Det-N pairs with {\it
this} and {\it that}, because its 3\% strong appearance in Det-N pairs with 
{\it these}, it is not a singular grammatical number KN.}
\label{fig:w_11_1835}
\end{figure}

The neuron $w^{11}_{1835}$ is an interesting case.  It appears as a
knowledge neuron in 92\% and 90\% of the determiner {\it this} and {\it that},
and only 3\% of the determiner {\it these}.  However, these 3\% of the neurons
are very strong.  Including $w^{11}_{1835}$ as a singular neuron, or using the
neuron on its own does not should good localisation of grammatical number 
(Figure \ref{fig:w_11_1835}).  Therefore, for our KN search, we excluded these
neurons.

\begin{table}[t]
\scriptsize\centering
\caption{Neurons identified using BLiMP's {determiner\_noun\_agreement\_2}
paradigm.}
\begin{tabular}{c|cccc} \toprule
Neuron                  & this & that & these & those \\ \midrule
{\red $w^{9}_{1094}$}   & 0.00 & 0.00 & 1.00  & 1.00  \\
$w^{11}_{1835}$         & 0.92 & 0.90 & 0.03  & 0.00  \\
{\blue $w^{10}_{2096}$} & 0.88 & 0.81 & 0.00  & 0.00  \\
$w^{11}_{2686}$         & 0.00 & 0.85 & 0.00  & 0.04  \\
$w^{9}_{2339}$          & 0.27 & 0.03 & 0.36  & 0.00  \\
$w^{10}_{2539}$         & 0.00 & 0.00 & 0.14  & 0.00  \\
$w^{10}_{2999}$         & 0.04 & 0.00 & 0.12  & 0.00  \\
$w^{11}_{10}$           & 0.01 & 0.00 & 0.07  & 0.03  \\
$w^{10}_{633}$          & 0.00 & 0.00 & 0.07  & 0.00  \\
$w^{10}_{651}$          & 0.00 & 0.00 & 0.07  & 0.09  \\
$w^{11}_{2231}$         & 0.07 & 0.00 & 0.00  & 0.00  \\
$w^{10}_{2029}$         & 0.00 & 0.00 & 0.06  & 0.06  \\
$w^{11}_{866}$          & 0.03 & 0.03 & 0.06  & 0.03  \\
$w^{10}_{141}$          & 0.00 & 0.00 & 0.04  & 0.01  \\
$w^{11}_{1405}$         & 0.04 & 0.01 & 0.00  & 0.00  \\
$w^{7}_{3034}$          & 0.00 & 0.00 & 0.03  & 0.00  \\
$w^{11}_{900}$          & 0.03 & 0.01 & 0.03  & 0.00  \\
$w^{10}_{2606}$         & 0.00 & 0.00 & 0.03  & 0.00  \\
$w^{11}_{1723}$         & 0.03 & 0.00 & 0.00  & 0.66  \\
$w^{10}_{35}$           & 0.03 & 0.01 & 0.00  & 0.01  \\
$w^{8}_{1404}$          & 0.03 & 0.00 & 0.00  & 0.00  \\
$w^{11}_{646}$          & 0.00 & 0.01 & 0.00  & 0.00  \\
$w^{10}_{1626}$         & 0.00 & 0.01 & 0.00  & 0.00  \\
$w^{9}_{412}$           & 0.00 & 0.01 & 0.00  & 0.00  \\
$w^{6}_{2123}$          & 0.00 & 0.00 & 0.01  & 0.00  \\
$w^{9}_{2766}$          & 0.00 & 0.00 & 0.01  & 0.01  \\
$w^{10}_{1845}$         & 0.00 & 0.00 & 0.01  & 0.00  \\
$w^{11}_{1248}$         & 0.00 & 0.00 & 0.01  & 0.00  \\
$w^{11}_{444}$          & 0.01 & 0.00 & 0.00  & 0.00  \\
$w^{7}_{1248}$          & 0.01 & 0.00 & 0.00  & 0.00  \\
$w^{11}_{2480}$         & 0.01 & 0.00 & 0.00  & 0.01  \\
$w^{11}_{1824}$         & 0.01 & 0.00 & 0.00  & 0.00  \\
$w^{8}_{2754}$          & 0.00 & 0.00 & 0.00  & 0.06  \\
$w^{10}_{606}$          & 0.00 & 0.00 & 0.00  & 0.01  \\
$w^{6}_{602}$           & 0.00 & 0.00 & 0.00  & 0.01  \\
$w^{10}_{175}$          & 0.00 & 0.00 & 0.00  & 0.10  \\
$w^{11}_{1568}$         & 0.00 & 0.00 & 0.00  & 0.01  \\
 \bottomrule
\end{tabular}
\label{tab:neurons_identified}
\end{table}

\newpage

\section{KNs for Linguistic Phenomena}
\label{app:kns_for_linguistic_phenomena}

In this section, we are going to present the KN search for other linguistic
phenomena and models.  Table \ref{tab:all_kns_1}, \ref{tab:all_kns_2} and \ref{tab:all_kns_3} lists
all the neurons we identified for each of the BLiMP paradigms and language
models.  Some of the paradigms such as determiner\_noun\_agreement\_2 are not
applicable for decoder only language models because the target precedes the
source.  We do not evaluate these paradigms for the two decoder-only language
models.

Similar to the search for determiner\_noun\_agreement\_2 KNs, we first identify
neurons with common appearance in the prompts with a certain grammatical
property.  Then, we manually test if suppressing these neurons can lead to
significant model behaviour change.  The number of neurons we identified is
different across model and paradigms.  The exact pattern behind this difference
may need more investigation.  We leave it for future work.

Please consult our code release repository%
\footnote{\href{https://github.com/frankniujc/kn\_thesis}{https://github.com/frankniujc/kn\_thesis}}
for any updates, detailed analysis and documentation of these KN search
results.

\begin{table}
\centering\tiny
\caption{BLiMP phenomena and paradigms.}
\begin{tabular}{p{1cm}p{2cm}clp{7cm}} \toprule
Phenomenon & Paradigms & Property & Model & KNs \\ \midrule
Anaphor Agreement & anaphor\_gender \_agreement & m & BERT
    & $w^{7}_{942}$, $w^{8}_{2881}$, $w^{10}_{1845}$ \\ \cmidrule{4-5}
& & & GPT-2 
    & $w^{9}_{2985}$, $w^{11}_{17}$, $w^{11}_{1611}$, $w^{11}_{2044}$, $w^{11}_{2910}$\\ \cmidrule{4-5}
& & & LLaMA-2
    & $w^{0}_{6454}$, $w^{30}_{5279}$, $w^{31}_{10638}$ \\ \cmidrule{3-5}
& & f & BERT
    &  $w^{7}_{942}$, $w^{9}_{1712}$ \\ \cmidrule{4-5}
& & & GPT-2 
    & $w^{0}_{1344}$, $w^{0}_{1403}$, $w^{8}_{1253}$, $w^{8}_{1891}$, $w^{10}_{2093}$ \\ \cmidrule{4-5}
& & & LLaMA-2
    & $w^{29}_{10442}$, $w^{30}_{3882}$, $w^{30}_{6935}$, $w^{31}_{2606}$, $w^{31}_{7984}$ \\ \cmidrule{2-5}
& anaphor\_number \_agreement & sg & BERT 
    & $w^{9}_{1712}$ \\ \cmidrule{4-5}
& & & GPT-2 
    & $w^{8}_{1891}$, $w^{10}_{1690}$ \\ \cmidrule{4-5}
& & & LLaMA-2
    & $w^{30}_{5279}$, $w^{31}_{7839}$, $w^{31}_{9148}$ \\ \cmidrule{3-5}
& & pl & BERT
    & $w^{11}_{2070}$ \\ \cmidrule{4-5}
& & & GPT-2 
    & $w^{10}_{3060}$, $w^{11}_{598}$ \\ \cmidrule{4-5}
& & & LLaMA-2
    & $w^{31}_{1116}$, $w^{31}_{6124}$, $w^{31}_{7742}$, $w^{31}_{8169}$ \\ \midrule
Determiner-Noun Agreement
& determiner\_noun \_agreement\_1 & sg & BERT
    & $w^{7}_{452}$, $w^{7}_{2631}$, $w^{8}_{1222}$, $w^{8}_{2660}$, $w^{9}_{620}$, $w^{9}_{1283}$, $w^{10}_{136}$, $w^{10}_{598}$, $w^{10}_{1038}$, $w^{10}_{1143}$, $w^{10}_{1279}$, $w^{10}_{1418}$, $w^{10}_{2162}$, $w^{11}_{384}$, $w^{11}_{526}$, $w^{11}_{558}$, $w^{11}_{762}$, $w^{11}_{870}$, $w^{11}_{991}$, $w^{11}_{999}$, $w^{11}_{1143}$, $w^{11}_{1267}$, $w^{11}_{1350}$, $w^{11}_{1435}$, $w^{11}_{1496}$, $w^{11}_{1521}$, $w^{11}_{1565}$, $w^{11}_{2204}$, $w^{11}_{2221}$, $w^{11}_{2592}$, $w^{11}_{2632}$, $w^{11}_{2772}$, $w^{11}_{2847}$, $w^{11}_{2994}$ \\ \cmidrule{4-5}
& & & GPT-2
    & $w^{0}_{11}$, $w^{0}_{37}$, $w^{0}_{132}$, $w^{0}_{1294}$, $w^{0}_{1311}$, $w^{0}_{1414}$, $w^{0}_{1529}$, $w^{0}_{1797}$, $w^{0}_{1950}$, $w^{0}_{2577}$, $w^{0}_{2733}$, $w^{7}_{2367}$, $w^{10}_{571}$, $w^{11}_{1998}$ \\ \cmidrule{4-5}
& & & LLaMA-2 
    & $w^{30}_{5279}$, $w^{30}_{8177}$, $w^{31}_{1876}$, $w^{31}_{5591}$, $w^{31}_{5876}$, $w^{31}_{8061}$, $w^{31}_{8236}$ \\ \cmidrule{3-5}
& & pl & BERT
    & $w^{8}_{52}$, $w^{9}_{218}$, $w^{9}_{698}$, $w^{9}_{1343}$, $w^{9}_{1812}$, $w^{9}_{2158}$, $w^{10}_{6}$, $w^{10}_{845}$, $w^{10}_{883}$, $w^{10}_{975}$, $w^{10}_{1178}$, $w^{10}_{1959}$, $w^{11}_{9}$, $w^{11}_{199}$, $w^{11}_{310}$, $w^{11}_{532}$, $w^{11}_{631}$, $w^{11}_{1009}$, $w^{11}_{1040}$, $w^{11}_{1396}$, $w^{11}_{1548}$, $w^{11}_{1767}$, $w^{11}_{1965}$, $w^{11}_{1985}$, $w^{11}_{2646}$, $w^{11}_{2978}$, $w^{11}_{2995}$ \\ \cmidrule{4-5}
& & & GPT-2
    & $w^{0}_{646}$, $w^{0}_{1013}$, $w^{0}_{1159}$, $w^{0}_{1227}$, $w^{0}_{1382}$, $w^{0}_{1469}$, $w^{0}_{1612}$, $w^{0}_{2428}$, $w^{0}_{2702}$, $w^{0}_{3055}$, $w^{7}_{1871}$, $w^{10}_{476}$, $w^{10}_{1693}$, $w^{10}_{3038}$, $w^{11}_{472}$, $w^{11}_{2387}$ \\ \cmidrule{4-5}
& & & LLaMA-2
    & $w^{2}_{7003}$, $w^{5}_{4435}$, $w^{15}_{6139}$, $w^{30}_{376}$, $w^{30}_{2262}$, $w^{30}_{3619}$, $w^{30}_{4228}$, $w^{30}_{4257}$, $w^{30}_{7935}$, $w^{30}_{9673}$ \\ \cmidrule{2-5}

& determiner\_noun & sg & BERT & $w_{2096}^{10}$ \\ \cmidrule{3-5}
& \_agreement\_2   & pl & BERT & $w_{1094}^{9}$  \\ \cmidrule{2-5}

& determiner\_noun \_agreement\_irregular\_1 & sg & BERT
    & $w^{7}_{655}$, $w^{8}_{656}$, $w^{8}_{1222}$, $w^{9}_{561}$, $w^{9}_{1146}$, $w^{9}_{2393}$, $w^{10}_{136}$, $w^{10}_{598}$, $w^{10}_{1143}$, $w^{10}_{1418}$, $w^{10}_{1955}$, $w^{11}_{237}$, $w^{11}_{281}$, $w^{11}_{326}$, $w^{11}_{558}$, $w^{11}_{1309}$, $w^{11}_{1350}$, $w^{11}_{1521}$, $w^{11}_{2257}$, $w^{11}_{2388}$, $w^{11}_{2592}$, $w^{11}_{2870}$, $w^{11}_{2994}$, $w^{11}_{3020}$ \\ \cmidrule{4-5}
& & & GPT-2
    & $w^{0}_{11}$, $w^{0}_{2845}$, $w^{6}_{628}$, $w^{6}_{2216}$, $w^{9}_{569}$, $w^{9}_{2603}$, $w^{10}_{383}$, $w^{10}_{534}$, $w^{10}_{571}$ \\ \cmidrule{4-5}
& & & LLaMA-2 
    & $w^{29}_{168}$, $w^{30}_{3619}$, $w^{30}_{3937}$, $w^{30}_{8298}$, $w^{31}_{5591}$, $w^{31}_{8169}$, $w^{31}_{10905}$ \\ \cmidrule{3-5}
& & pl & BERT 
    & $w^{8}_{2594}$, $w^{9}_{698}$, $w^{9}_{1106}$, $w^{9}_{2158}$, $w^{10}_{664}$, $w^{10}_{845}$, $w^{10}_{1178}$, $w^{10}_{1547}$, $w^{10}_{2810}$, $w^{11}_{26}$, $w^{11}_{45}$, $w^{11}_{310}$, $w^{11}_{532}$, $w^{11}_{631}$, $w^{11}_{1239}$, $w^{11}_{1873}$, $w^{11}_{1934}$, $w^{11}_{1965}$, $w^{11}_{2070}$, $w^{11}_{2320}$, $w^{11}_{2944}$, $w^{11}_{2978}$, $w^{11}_{2995}$ \\ \cmidrule{4-5}
& & & GPT-2
    & $w^{0}_{185}$, $w^{0}_{280}$, $w^{7}_{1871}$, $w^{8}_{55}$, $w^{9}_{792}$, $w^{10}_{1693}$, $w^{10}_{3038}$, $w^{11}_{472}$, $w^{11}_{2387}$ \\ \cmidrule{4-5}
& & & LLaMA-2
    & $w^{30}_{1443}$, $w^{30}_{1686}$, $w^{30}_{3397}$, $w^{30}_{7455}$, $w^{30}_{8343}$, $w^{30}_{8878}$, $w^{30}_{9673}$, $w^{31}_{6587}$ \\ \cmidrule{2-5}
& determiner\_noun  & sg & BERT & $w_{2096}^{10}$ \\ \cmidrule{3-5}
& \_agreement\_irregular\_2 & pl & BERT & $w_{1094}^{9}$ \\ %\cmidrule{2-5}

\bottomrule
\end{tabular}
\label{tab:all_kns_1}
\end{table}

\begin{table}
\centering\tiny
\caption{BLiMP phenomena and paradigms (continued).}
\begin{tabular}{p{1cm}p{2cm}clp{7cm}} \toprule
Phenomenon & Paradigms & Property & Model & KNs \\ \midrule
Determiner-Noun Agreement (continued)

& determiner\_noun \_agreement\_with\_adj\_1 & sg & BERT
    & $w^{7}_{64}$, $w^{7}_{984}$, $w^{9}_{391}$, $w^{9}_{1283}$, $w^{9}_{2381}$, $w^{9}_{2951}$, $w^{10}_{995}$, $w^{10}_{1235}$, $w^{10}_{1269}$, $w^{10}_{1279}$, $w^{10}_{1382}$, $w^{10}_{1737}$, $w^{10}_{1955}$, $w^{10}_{2024}$, $w^{10}_{2935}$, $w^{11}_{74}$, $w^{11}_{558}$, $w^{11}_{991}$, $w^{11}_{1350}$, $w^{11}_{1521}$, $w^{11}_{2173}$, $w^{11}_{2647}$ \\ \cmidrule{4-5}
& & & GPT-2
    & $w^{0}_{1344}$, $w^{5}_{1888}$, $w^{7}_{1871}$, $w^{8}_{1253}$, $w^{9}_{330}$, $w^{10}_{80}$, $w^{10}_{379}$, $w^{10}_{383}$, $w^{10}_{2178}$, $w^{10}_{2738}$, $w^{11}_{740}$, $w^{11}_{1593}$, $w^{11}_{2044}$ \\ \cmidrule{4-5}
& & & LLaMA-2
    & $w^{30}_{5279}$, $w^{30}_{8177}$, $w^{31}_{1876}$, $w^{31}_{5591}$, $w^{31}_{5876}$, $w^{31}_{8061}$, $w^{31}_{8236}$ \\ \cmidrule{3-5}
& & pl & BERT
    & $w^{8}_{3052}$, $w^{9}_{698}$, $w^{9}_{1343}$, $w^{9}_{1812}$, $w^{9}_{2158}$, $w^{9}_{2327}$, $w^{10}_{6}$, $w^{10}_{664}$, $w^{10}_{845}$, $w^{10}_{1178}$, $w^{10}_{1624}$, $w^{10}_{1888}$, $w^{10}_{1959}$, $w^{11}_{122}$, $w^{11}_{291}$, $w^{11}_{310}$, $w^{11}_{532}$, $w^{11}_{1009}$, $w^{11}_{2042}$, $w^{11}_{2070}$, $w^{11}_{2106}$, $w^{11}_{2978}$, $w^{11}_{2995}$ \\ \cmidrule{4-5}
& & & GPT-2
    & $w^{3}_{289}$, $w^{8}_{1993}$, $w^{9}_{840}$, $w^{10}_{54}$, $w^{10}_{476}$, $w^{10}_{1693}$, $w^{10}_{3038}$, $w^{11}_{713}$, $w^{11}_{992}$, $w^{11}_{2387}$, $w^{11}_{2408}$, $w^{11}_{2605}$ \\ \cmidrule{4-5}
& & & LLaMA-2
    & $w^{2}_{7003}$, $w^{5}_{4435}$, $w^{15}_{6139}$, $w^{30}_{376}$, $w^{30}_{2262}$, $w^{30}_{3619}$, $w^{30}_{4228}$, $w^{30}_{4257}$, $w^{30}_{7935}$, $w^{30}_{9673}$ \\ \cmidrule{2-5}

& determiner\_noun  & sg  & BERT
    & $w_{2096}^{10}$ \\ \cmidrule{3-5}
& \_agreement\_with\_adj\_2 & pl & BERT
    & $w_{1094}^{9}$ \\ \cmidrule{2-5}

& determiner\_noun \_agreement\_with \_adj\_irregular\_1 & sg & BERT
    & $w^{8}_{1985}$, $w^{9}_{391}$, $w^{9}_{1450}$, $w^{9}_{2407}$, $w^{10}_{136}$, $w^{10}_{247}$, $w^{10}_{598}$, $w^{10}_{694}$, $w^{10}_{1143}$, $w^{10}_{1955}$, $w^{10}_{2024}$, $w^{11}_{155}$, $w^{11}_{281}$, $w^{11}_{558}$, $w^{11}_{1084}$, $w^{11}_{1224}$, $w^{11}_{1315}$, $w^{11}_{1350}$, $w^{11}_{1521}$, $w^{11}_{1938}$, $w^{11}_{3070}$ \\ \cmidrule{4-5}
& & & GPT-2
    & $w^{0}_{1344}$, $w^{5}_{1888}$, $w^{7}_{1871}$, $w^{8}_{1253}$, $w^{9}_{330}$, $w^{10}_{80}$, $w^{10}_{379}$, $w^{10}_{383}$, $w^{10}_{2178}$, $w^{10}_{2738}$, $w^{11}_{740}$, $w^{11}_{1593}$, $w^{11}_{2044}$ \\ \cmidrule{4-5}
& & & LLaMA-2
    & $w^{2}_{7003}$, $w^{30}_{3937}$, $w^{30}_{6935}$, $w^{30}_{8298}$, $w^{30}_{9131}$, $w^{31}_{7988}$, $w^{31}_{8236}$ \\ \cmidrule{3-5}
& & pl & BERT
    & $w^{8}_{426}$, $w^{9}_{698}$, $w^{9}_{2158}$, $w^{10}_{845}$, $w^{10}_{933}$, $w^{10}_{1178}$, $w^{10}_{1911}$, $w^{11}_{165}$, $w^{11}_{211}$, $w^{11}_{307}$, $w^{11}_{310}$, $w^{11}_{532}$, $w^{11}_{662}$, $w^{11}_{1107}$, $w^{11}_{1177}$, $w^{11}_{1548}$, $w^{11}_{1934}$, $w^{11}_{1965}$, $w^{11}_{2070}$, $w^{11}_{2106}$, $w^{11}_{2621}$, $w^{11}_{2704}$, $w^{11}_{2944}$, $w^{11}_{2995}$, $w^{11}_{3037}$ \\ \cmidrule{4-5}
& & & GPT-2
    & $w^{0}_{2702}$, $w^{8}_{1993}$, $w^{10}_{54}$, $w^{10}_{900}$, $w^{10}_{1693}$, $w^{10}_{3038}$, $w^{11}_{472}$, $w^{11}_{2387}$, $w^{11}_{2408}$, $w^{11}_{2605}$ \\ \cmidrule{4-5}
& & & LLaMA-2
    & $w^{15}_{5180}$, $w^{30}_{8878}$, $w^{30}_{10417}$, $w^{30}_{10552}$, $w^{30}_{10588}$, $w^{31}_{6587}$, $w^{31}_{11000}$ \\ \cmidrule{2-5}

& determiner\_noun                      &  sg       & BERT
    & $w_{2096}^{10}$ \\ \cmidrule{3-5}
& \_agreement\_with \_adj\_irregular\_2 &  pl       & BERT
    & $w_{1094}^{9}$ \\ \midrule
Subject- Verb Agreement
& regular\_plural\_subject \_verb\_agreement\_1 & sg & BERT 
    & $w^{10}_{455}$, $w^{11}_{153}$, $w^{11}_{1541}$ \\ \cmidrule{4-5}
& & & GPT-2
    & $w^{10}_{379}$, $w^{10}_{729}$, $w^{10}_{2839}$, $w^{11}_{2173}$, $w^{11}_{2187}$ \\ \cmidrule{4-5}
& & & LLaMA-2
    &  $w^{31}_{566}$ \\ \cmidrule{3-5}

& & pl & BERT
    & $w^{11}_{1073}$, $w^{11}_{1307}$ \\ \cmidrule{4-5}
& & & GPT-2
    & $w^{0}_{101}$, $w^{6}_{2674}$, $w^{10}_{1318}$, $w^{10}_{1347}$, $w^{10}_{2409}$, $w^{11}_{17}$, $w^{11}_{627}$, $w^{11}_{896}$, $w^{11}_{1043}$, $w^{11}_{3051}$
 \\ \cmidrule{4-5}
& & & LLaMA-2
    &  $w^{17}_{4054}$, $w^{30}_{5279}$, $w^{31}_{556}$ \\ \cmidrule{2-5}

& regular\_plural\_subject \_verb\_agreement\_2 & sg & BERT 
    & $w^{11}_{1350}$ \\ \cmidrule{3-5}
& & pl & BERT
    & $w^{10}_{1178}$, $w^{11}_{2995}$ \\ \cmidrule{2-5}

& irregular\_plural\_subject \_verb\_agreement\_1 & sg & BERT
    & $w^{10}_{455}$, $w^{11}_{153}$, $w^{11}_{1541}$ \\ \cmidrule{4-5}
& & & GPT-2
    & $w^{10}_{379}$, $w^{10}_{729}$, $w^{10}_{2839}$, $w^{11}_{2173}$, $w^{11}_{2187}$ \\ \cmidrule{4-5}
& & & LLaMA-2
    &  $w^{31}_{566}$ \\ \cmidrule{3-5}
& & pl & BERT
    & $w^{11}_{1073}$, $w^{11}_{1307}$ \\ \cmidrule{4-5}
& & & GPT-2
    & $w^{0}_{101}$, $w^{6}_{2674}$, $w^{10}_{1318}$, $w^{10}_{1347}$, $w^{10}_{2409}$, $w^{11}_{17}$, $w^{11}_{627}$, $w^{11}_{896}$, $w^{11}_{1043}$, $w^{11}_{3051}$ \\ \cmidrule{4-5}
& & & LLaMA-2
    &  $w^{30}_{5279}$, $w^{31}_{212}$, $w^{31}_{2606}$, $w^{31}_{7839}$ \\ \cmidrule{2-5}

& irregular\_plural\_subject \_verb\_agreement\_2 & sg & BERT 
    & $w^{11}_{1350}$ \\ \cmidrule{3-5}
& & pl & BERT
    & $w^{10}_{1178}$, $w^{11}_{2995}$ \\
\bottomrule
\end{tabular}
\label{tab:all_kns_2}
\end{table}

\begin{table}
\centering\tiny
\caption{BLiMP phenomena and paradigms (continued).}
\begin{tabular}{p{1cm}p{2cm}clp{7cm}} \toprule
Phenomenon & Paradigms & Property & Model & KNs \\ \midrule
Subject-Verb Agreement (continued)

& distractor\_agreement \_relational\_noun & sg & BERT
    & $w^{10}_{455}$, $w^{11}_{153}$, $w^{11}_{1541}$ \\ \cmidrule{4-5}
& & & GPT-2
    & $w^{10}_{379}$, $w^{10}_{729}$, $w^{10}_{2839}$, $w^{11}_{2173}$, $w^{11}_{2187}$ \\ \cmidrule{4-5}
& & & LLaMA-2
    &  $w^{31}_{566}$ \\ \cmidrule{3-5}
& & pl & BERT
    & $w^{9}_{2253}$, $w^{11}_{1307}$ \\ \cmidrule{4-5}
& & & GPT-2
    & $w^{0}_{101}$, $w^{6}_{2674}$, $w^{10}_{1318}$, $w^{10}_{1347}$, $w^{10}_{2409}$, $w^{11}_{17}$, $w^{11}_{627}$, $w^{11}_{896}$, $w^{11}_{1043}$, $w^{11}_{3051}$ \\ \cmidrule{4-5}
& & & LLaMA-2
    &  $w^{30}_{5279}$, $w^{31}_{3336}$, $w^{31}_{3658}$, $w^{31}_{7342}$ \\ \cmidrule{2-5}

& distractor\_agreement \_relative\_clause & sg & BERT
    & $w^{11}_{798}$, $w^{11}_{1541}$ \\ \cmidrule{4-5}
& & & GPT-2
    & $w^{10}_{379}$, $w^{10}_{729}$, $w^{10}_{2839}$, $w^{11}_{2173}$, $w^{11}_{2187}$ \\ \cmidrule{4-5}
& & & LLaMA-2
    &  $w^{31}_{566}$  \\ \cmidrule{3-5}
& & pl & BERT
    & $w^{9}_{2253}$ \\ \cmidrule{4-5}
& & & GPT-2
    & $w^{0}_{101}$, $w^{6}_{2674}$, $w^{10}_{1318}$, $w^{10}_{1347}$, $w^{10}_{2409}$, $w^{11}_{17}$, $w^{11}_{627}$, $w^{11}_{896}$, $w^{11}_{1043}$, $w^{11}_{3051}$ \\ \cmidrule{4-5}
& & & LLaMA-2
    &  $w^{30}_{5279}$, $w^{31}_{212}$, $w^{31}_{2606}$, $w^{31}_{7839}$ \\
\bottomrule
\end{tabular}
\label{tab:all_kns_3}
\end{table}

\newpage

\subsection{Effects of Suppressing the KNs}
\label{sub:effects_of_suppressing_the_kns}

\begin{figure}

\begin{subfigure}[b]{\linewidth}
\includegraphics[width=0.5\linewidth]
{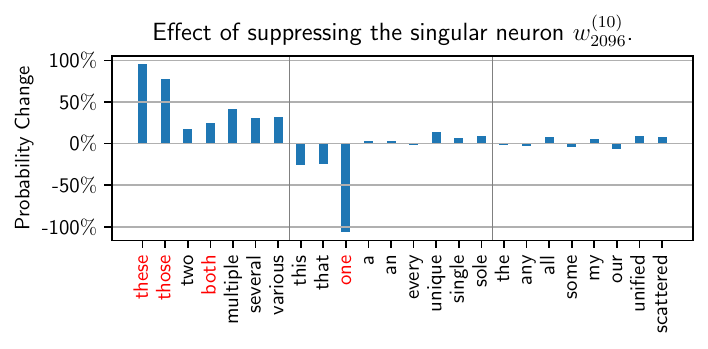} \hfill
\includegraphics[width=0.5\linewidth]
{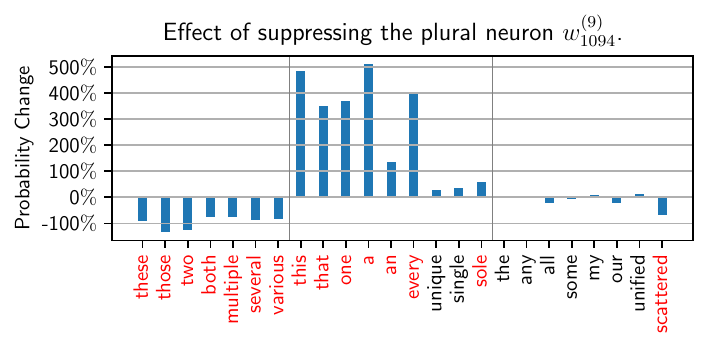}
\caption{Base case Det-N agreement: {determiner\_noun\_agreement\_2}.}
\end{subfigure}

\begin{subfigure}[b]{\linewidth}
\includegraphics[width=0.5\linewidth]
{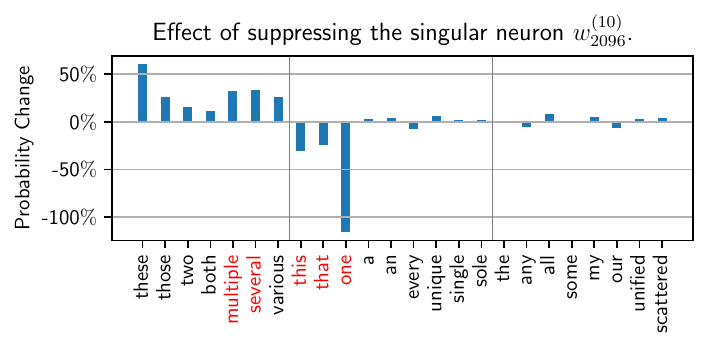} \hfill
\includegraphics[width=0.5\linewidth]
{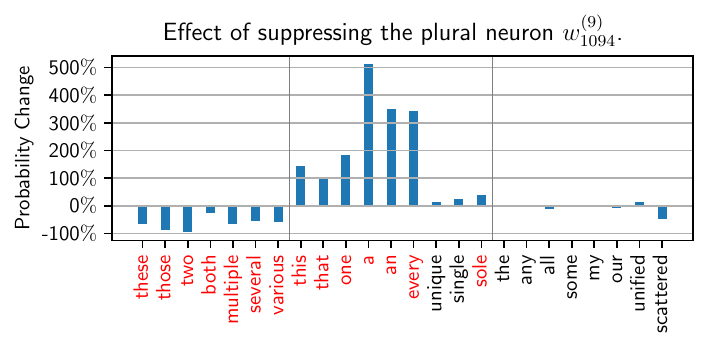}
\caption{Det-N agreement with irregular noun inflections:
{determiner\_noun\_agreement\_with\_adj\_2}.}
\end{subfigure}

\begin{subfigure}[b]{\linewidth}
\includegraphics[width=0.5\linewidth]
{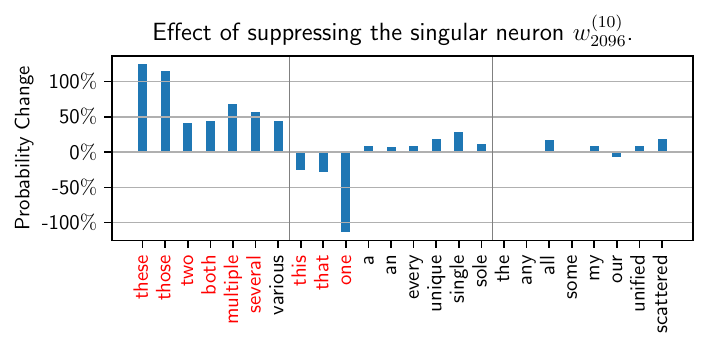} \hfill
\includegraphics[width=0.5\linewidth]
{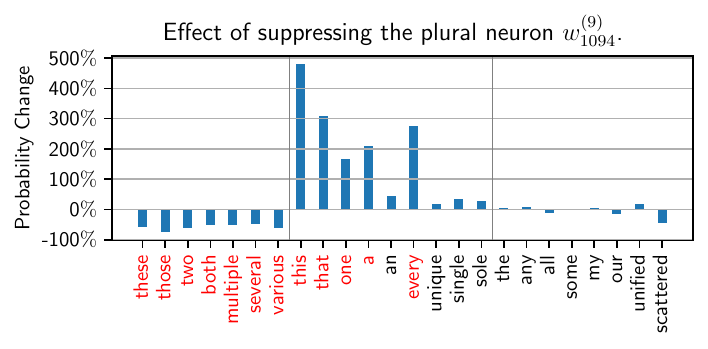}
\caption{Det-N agreement with an adjective distractor:
{determiner\_noun\_agreement\_irregular\_2}.}
\end{subfigure}

\begin{subfigure}[b]{\linewidth}
\includegraphics[width=0.5\linewidth]
{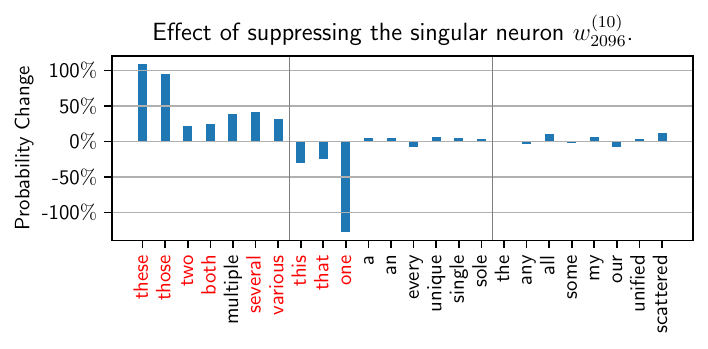} \hfill
\includegraphics[width=0.5\linewidth]
{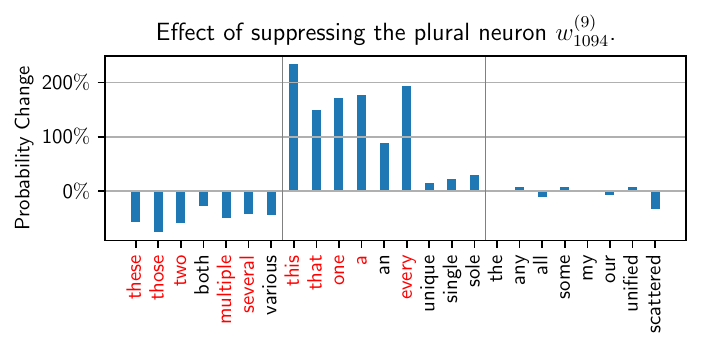}
\caption{Adjective distractor and irregular nouns:
{determiner\_noun\_agreement\_with\_adj\_irregular\_2}}
\end{subfigure}
\caption{Effect of suppressing the KNs of type 2 determiner-noun agreement on
BERT.  We can observe similar levels of probability change as
determiner\_noun\_agreement\_2 with or without adding distractors (adjectives
and irregular verbs).}
\label{fig:det_n_agr_2}
\end{figure}

Figure \ref{fig:det_n_agr_2} shows the probability change after erasing the
identified singular or plural neuron.  All the results are similar to the
base case paradigm presented in the main section of the paper. We can
observe similar levels of probability change as determiner\_noun\_agreement\_2
with or without adding distractors (adjectives and irregular verbs).

\begin{figure}
\begin{subfigure}[b]{\linewidth}
\includegraphics[width=0.5\linewidth]
{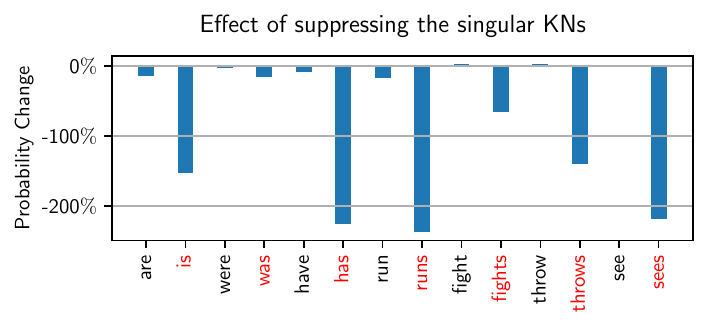} \hfill
\includegraphics[width=0.5\linewidth]
{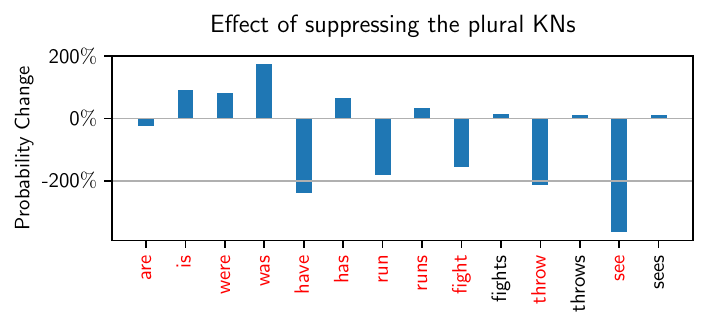}
\caption{Base case subject-verb agreement: {regular\_plural\_subject\_verb\_agreement\_1}.}
\end{subfigure}

\begin{subfigure}[b]{\linewidth}
\includegraphics[width=0.5\linewidth]
{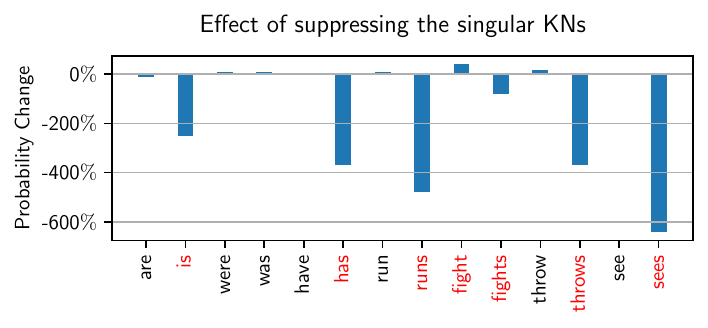} \hfill
\includegraphics[width=0.5\linewidth]
{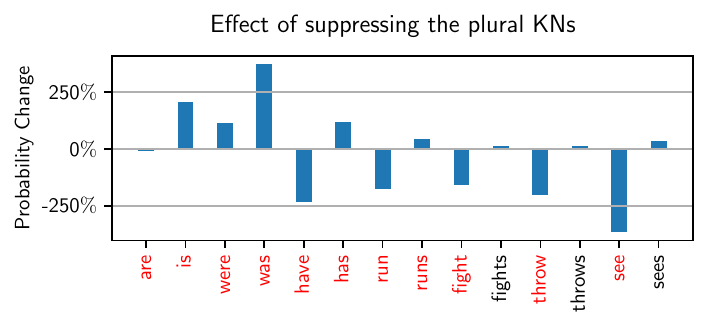}
\caption{Subject-verb agreement with irregular verbs: {irregular\_plural\_subject\_verb\_agreement\_1}.}
\end{subfigure}

\begin{subfigure}[b]{\linewidth}
\includegraphics[width=0.5\linewidth]
{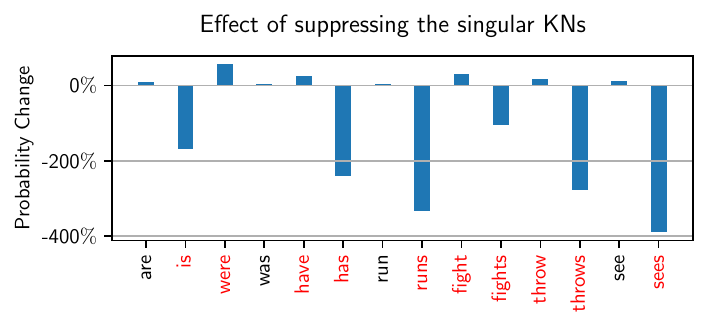} \hfill
\includegraphics[width=0.5\linewidth]
{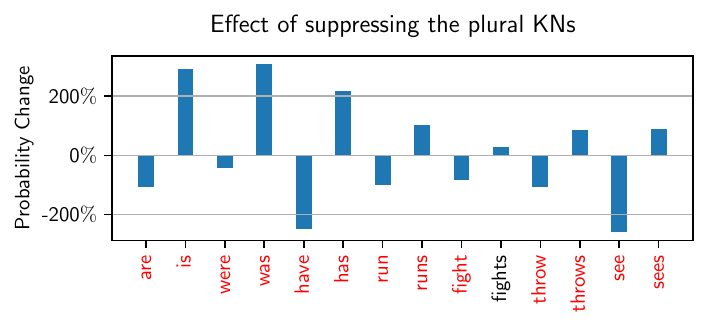}
\caption{Subject-verb agreement with relational noun distractor: {distractor\_agreement\_relational\_noun}.}
\end{subfigure}

\begin{subfigure}[b]{\linewidth}
\includegraphics[width=0.5\linewidth]
{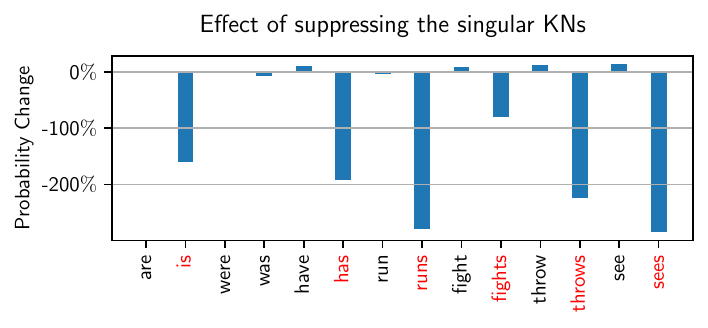} \hfill
\includegraphics[width=0.5\linewidth]
{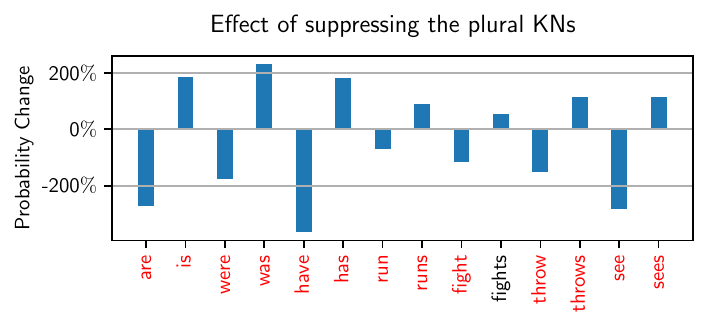}
\caption{Subject-verb agreement with relative clause distractor: {distractor\_agreement\_relative\_clause}.}
\end{subfigure}

\begin{subfigure}[b]{\linewidth}
\includegraphics[width=0.5\linewidth]
{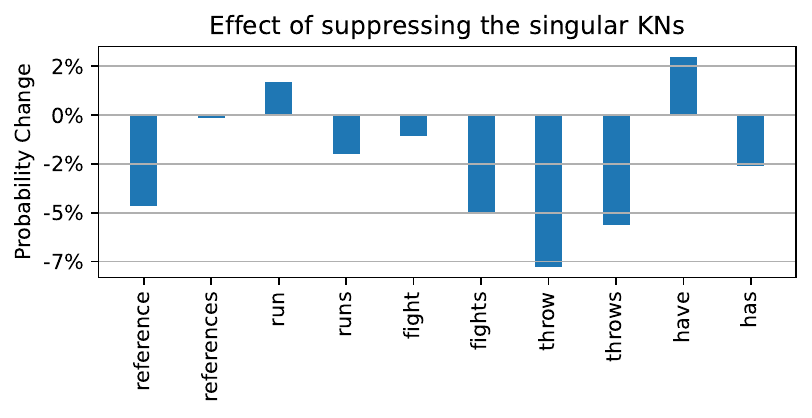} \hfill
\includegraphics[width=0.5\linewidth]
{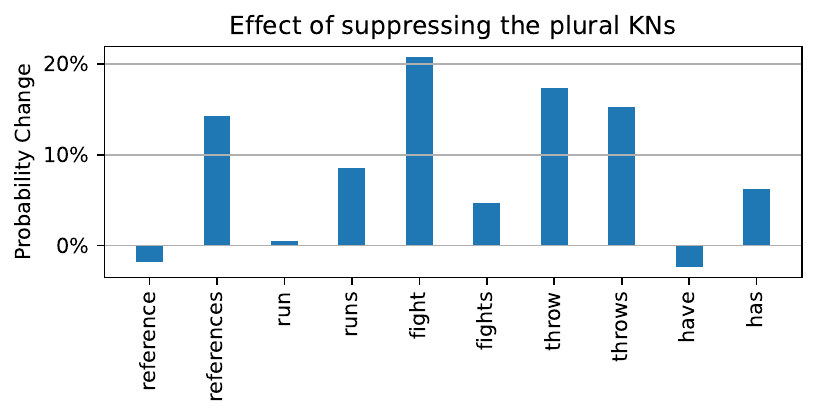}
\caption{LLaMA-2 results on {regular\_plural\_subject\_verb\_agreement\_1}. 
The effect of editing is less pronounced on larger LMs, as each neuron has
smaller relative importance.}
\end{subfigure}

\caption{Effect of suppressing the subject-verb agreement KNs on GPT-2.}
\label{fig:gpt2-sva}
\end{figure}

Figure \ref{fig:gpt2-sva} shows the probability change of subject-verb agreement
paradigms on GPT-2 and LLaMA-2.  For GPT-2, we can see that the suppression of
the singular neurons causes significant probability decrease for singular verbs
and essentially no effect to plural verbs.  The effect of intervention is more
pronounced for the plural neurons.  For LLaMA-2, however, as the model becomes
larger, the relative importance of each neuron becomes smaller.  Therefore, we
observe that the effect of KN editing is less pronounced.

\newpage

\subsection{Layer Distribution of Identified KNs}
\label{app:layer_dist_more}

\begin{figure}
\centering
\begin{subfigure}[b]{\linewidth}
\centering
\includegraphics[width=0.245\linewidth]{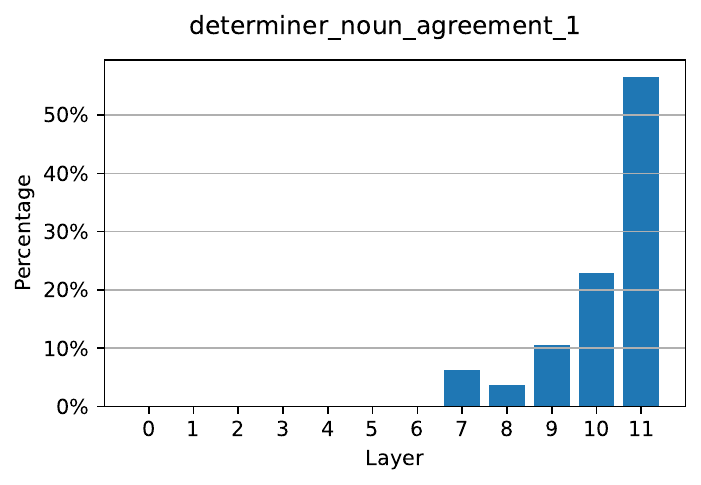}
\includegraphics[width=0.245\linewidth]{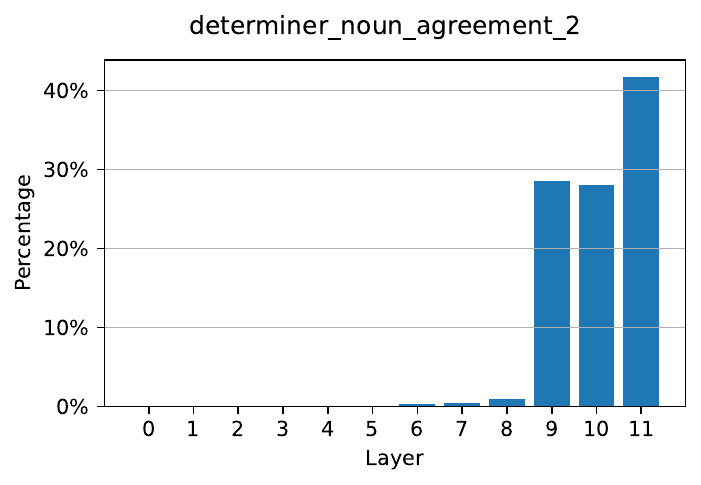}
\includegraphics[width=0.245\linewidth]{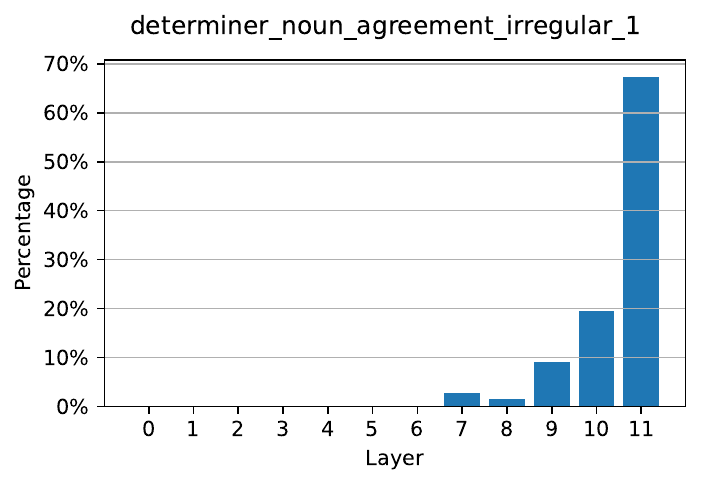}
\includegraphics[width=0.245\linewidth]{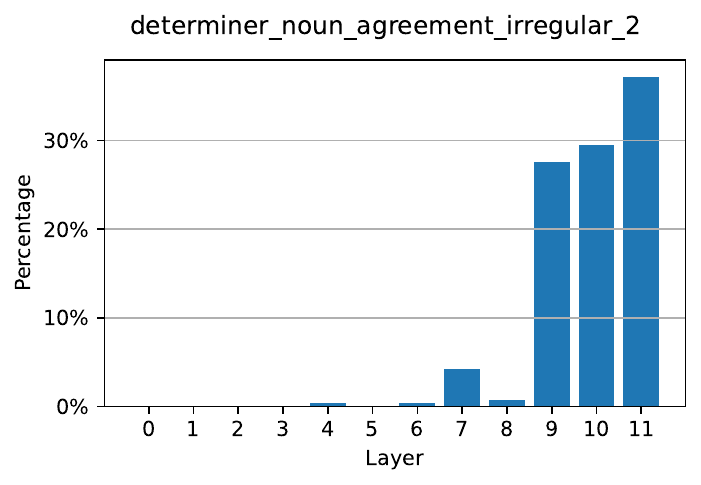}
\includegraphics[width=0.245\linewidth]{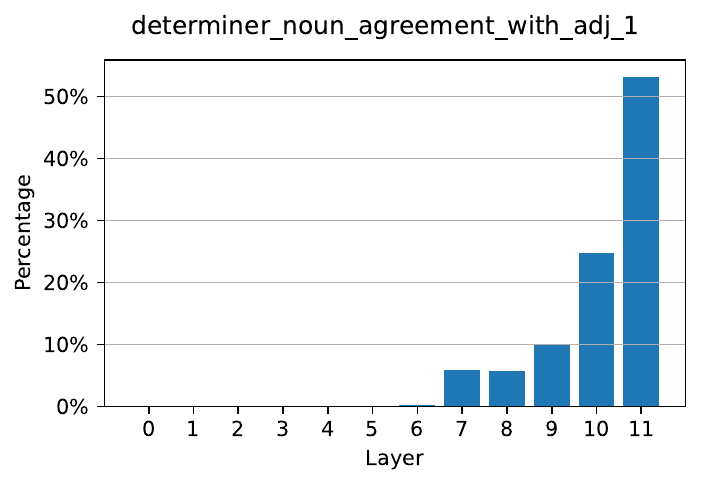}
\includegraphics[width=0.245\linewidth]{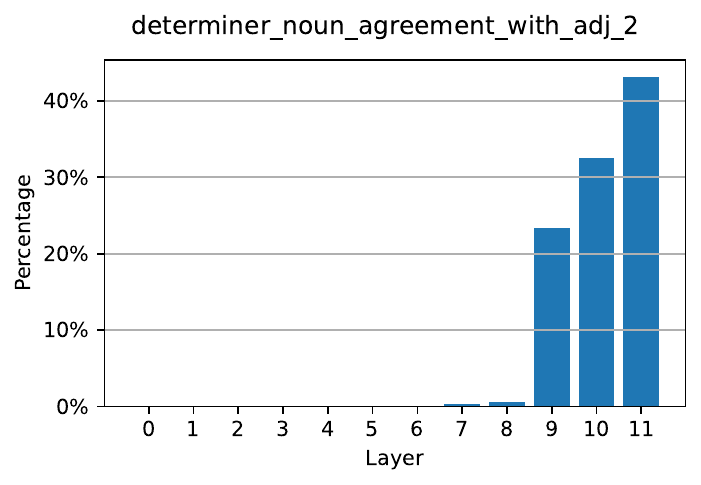}
\includegraphics[width=0.245\linewidth]{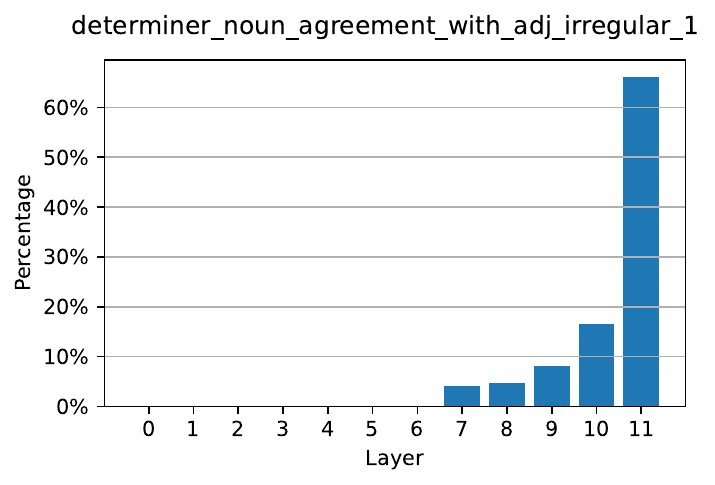}
\includegraphics[width=0.245\linewidth]{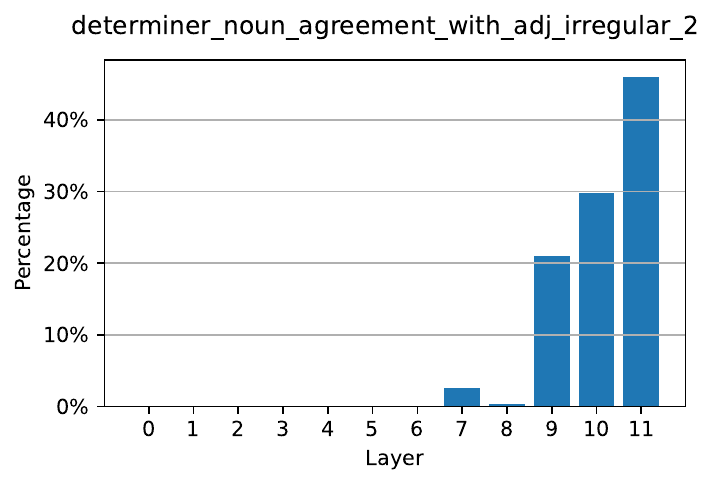}
\caption{Layer distribution of BLiMP determiner-noun agreement paradigms.}
\end{subfigure}

\begin{subfigure}[b]{\linewidth}
\centering
\includegraphics[width=0.245\linewidth]{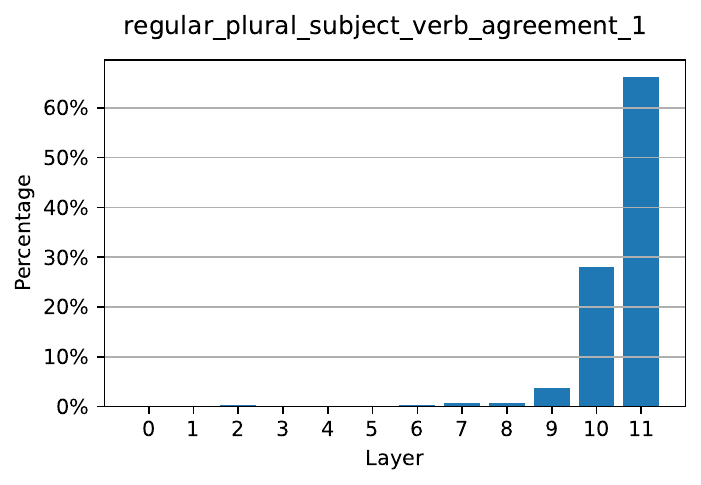}
\includegraphics[width=0.245\linewidth]{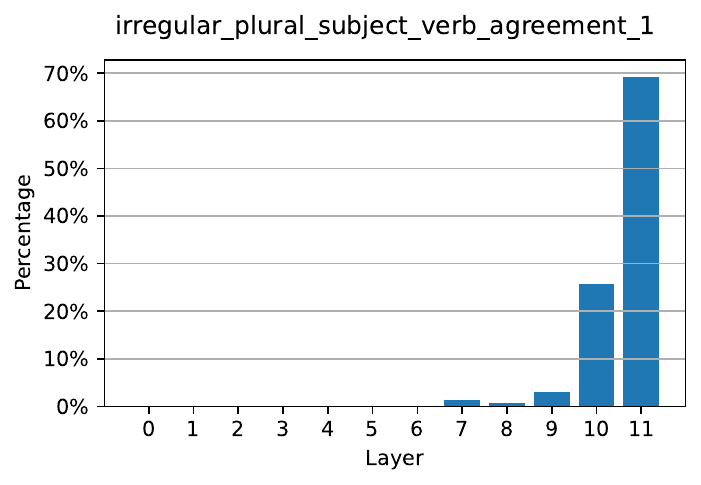}
\includegraphics[width=0.245\linewidth]{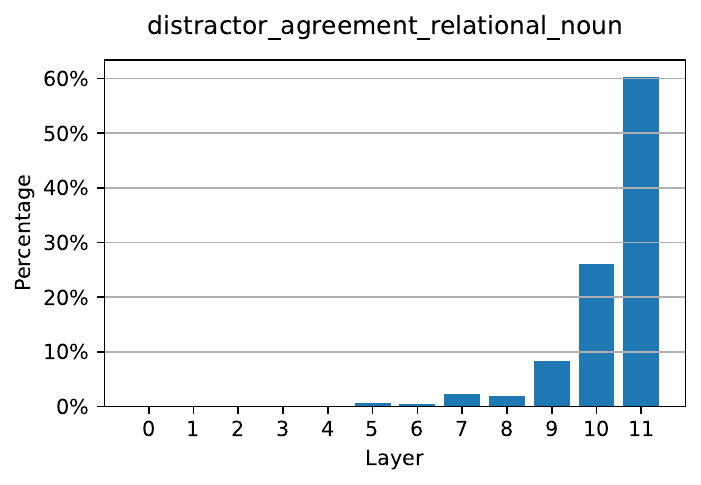}
\includegraphics[width=0.245\linewidth]{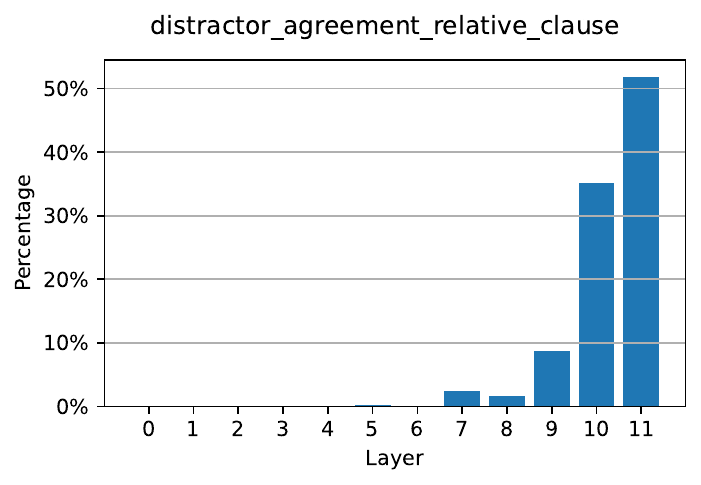}
\caption{Layer distribution of BLiMP subject-verb agreement paradigms.}
\end{subfigure}

\begin{subfigure}[b]{\linewidth}
\centering
\includegraphics[width=0.245\linewidth]{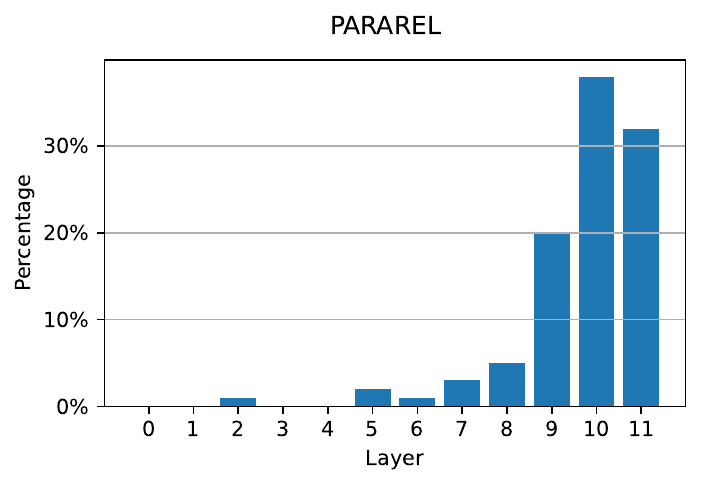}
\caption{Recreation of \cites{daiKnowledgeNeuronsPretrained2022} KN
distribution result across ParaRel relations (Figure 3).}
\end{subfigure}

\caption{Percentage of knowledge neurons identified in different BLiMP paradigms
using BERT.}
\label{fig:all_layers}
\end{figure}

In Figure \ref{fig:all_layers}, we analysed the layers of the identified KNs on
each BLiMP paradigm
and compare
it with \cites{daiKnowledgeNeuronsPretrained2022} finding on {\sc ParaRel}.
We notice that the vast
majority of neurons of all types are
distributed in the topmost layers of BERT.  This result confirms 
\cites{daiKnowledgeNeuronsPretrained2022} observation but disproved their
position.  There is nothing unique to fact-related neurons.  This finding agrees
with \cites{niuDoesBERTRediscover2022} refutation of 
\citet{jawaharWhatDoesBERT2019} and \citet{tenneyBERTRediscoversClassical2019}.
Syntactic information (formal competence) and semantic information (functional
competence) do not occupy different layers of the language model.  

\subsection{Levels of localisation}
\label{sub:level_of_localisation}

Table \ref{tab:levels_of_localisation} shows the levels of localisation on BERT
across all determiner-noun agreement (DNA) BLiMP paradigms.  These measures are
comparable to {\sc ParaRel} results shown in Table 
\ref{tab:levels_of_localisation_pararel}.

\begin{table}[ht]
\centering\scriptsize
\caption{Levels of localisation of determiner-noun agreement.}
\begin{tabular}{l|c|c|c|c|c|c|c|c|c} \toprule
\multirow{2}{*}{Paradigm} & \multicolumn{3}{c|}{BERT} & \multicolumn{3}{c|}{GPT-2} & \multicolumn{3}{c}{LLaMA-2} \\
 & $|\text{KN}|$ & $\tau$  & $R_1^2$ & $|\text{KN}|$ & $\tau$  & $R_1^2$ & $|\text{KN}|$ & $\tau$  & $R_1^2$ \\ \midrule
determiner\_noun\_agreement\_1                       & 3.94 & 0.71 & 0.56 &  0.06 & 0.45 & 0.15 & 3.38 & 0.44 & 0.24 \\
determiner\_noun\_agreement\_2                       & 1.86 & 0.62 & 0.56 &  -    & -    & -    & -    & -    & -    \\
determiner\_noun\_agreement\_irregular\_1            & 5.53 & 0.73 & 0.64 &  1.32 & 0.58 & 0.24 & 4.93 & 0.48 & 0.46 \\
determiner\_noun\_agreement\_irregular\_2            & 2.45 & 0.67 & 0.55 &  -    & -    & -    & -    & -    & -    \\
determiner\_noun\_agreement\_with\_adjective\_1      & 8.88 & 0.78 & 0.67 &  1.31 & 0.62 & 0.17 & 3.38 & 0.44 & 0.24 \\
determiner\_noun\_agreement\_with\_adjective\_2      & 2.26 & 0.67 & 0.57 &  -    & -    & -    & -    & -    & -    \\
determiner\_noun\_agreement\_with\_adj\_irregular\_1 & 9.79 & 0.78 & 0.67 &  0.12 & 0.51 & 0.15 & 4.55 & 0.45 & 0.48 \\
determiner\_noun\_agreement\_with\_adj\_irregular\_2 & 2.60 & 0.69 & 0.58 &  -  
& -    & -    & -    & -    & -    \\
% regular\_plural\_subject\_verb\_agreement\_1         & 2.00 & 0.70 & 0.51 & & 0.33 & 0.46 & 0.24 \\
% irregular\_plural\_subject\_verb\_agreement\_1       & 2.00 & 0.70 & 0.32 &  \\
% distractor\_agreement\_relational\_noun              & 3.00 & 0.70 & 0.83 &  \\
% distractor\_agreement\_relative\_clause              & 4.00 & 0.70 & 0.52 &  \\
\bottomrule
\end{tabular}
\label{tab:levels_of_localisation}
\end{table}

\begin{table}[ht]
\caption{Levels of localisation of different {\sc ParaRel} relations.}
\small\centering
\begin{tabular}{l|ccc|ccc|ccc} \toprule
\multirow{2}{*}{Relation} &
\multicolumn{3}{c|}{BERT} &
\multicolumn{3}{c|}{GPT-2} &
\multicolumn{3}{c}{LLaMA-2} \\
      & $|$KN$|$ & $\tau$   & $R_1^2$ & $|$KN$|$ & $\tau$   & $R_1^2$ & $|$KN$|$ & $\tau$   & $R_1^2$ \\ \midrule
P101  & 0.167    & 0.515 & 0.399   & 1.537    & 0.708 & 0.278   & 1.0      & 0.61  & 0.306   \\
P103  & 0.204    & 0.662 & 0.399   & 1.968    & 0.649 & 0.375   & 7.53     & 0.733 & 0.410   \\
P106  & 1.292    & 0.607 & 0.365   & 10.090   & 0.853 & 0.599   & 0.28     & 0.438 & 0.258   \\
P108  & 1.493    & 0.663 & 0.473   & 10.433   & 0.848 & 0.269   & 18.1     & 0.735 & 0.599   \\
P127  & 1.512    & 0.630 & 0.552   & 11.758   & 0.769 & 0.549   & 0.3      & 0.585 & 0.163   \\
P1303 & 10.462   & 0.814 & 0.684   & 12.453   & 0.771 & 0.573   & 0.3      & 0.63  & 0.303   \\
P136  & 14.862   & 0.856 & 0.646   & 14.435   & 0.878 & 0.677   & 2.1      & 0.64  & 0.754   \\
P1376 & 15.640   & 0.842 & 0.628   & 14.892   & 0.794 & 0.624   & 1.4      & 0.592 & 0.435   \\
P138  & 16.992   & 0.958 & 0.874   & 15.365   & 0.794 & 0.621   & 1.4      & 0.62  & 0.605   \\
P140  & 2.008    & 0.689 & 0.263   & 16.543   & 0.848 & 0.707   & 0.7      & 0.64  & 0.290   \\
P1412 & 2.196    & 0.687 & 0.612   & 18.286   & 0.782 & 0.618   & 1.8      & 0.66  & 0.590   \\
P159  & 2.200    & 0.666 & 0.392   & 19.285   & 0.838 & 0.626   & 9.5      & 0.765 & 0.637   \\
P176  & 2.376    & 0.680 & 0.254   & 19.299   & 0.823 & 0.650   & 2.7      & 0.675 & 0.524   \\
P178  & 2.553    & 0.686 & 0.392   & 19.898   & 0.828 & 0.686   & 2.1      & 0.65  & 0.414   \\
P19   & 2.597    & 0.693 & 0.481   & 2.603    & 0.661 & 0.417   & 0.24     & 0.482 & 0.526   \\
P190  & 2.920    & 0.673 & 0.308   & 21.989   & 0.949 & 0.757   & 0.8      & 0.54  & 0.216   \\
P20   & 28.993   & 0.883 & 0.644   & 22.984   & 0.871 & 0.671   & 0.52     & 0.458 & 0.377   \\
P264  & 3.335    & 0.667 & 0.213   & 23.082   & 0.867 & 0.685   & 1.7      & 0.605 & 0.416   \\
P27   & 3.925    & 0.722 & 0.483   & 29.296   & 0.862 & 0.619   & 1.8      & 0.67  & 0.376   \\
P279  & 4.286    & 0.691 & 0.657   & 3.437    & 0.649 & 0.360   & 14.7     & 0.755 & 0.683   \\
P30   & 4.338    & 0.698 & 0.449   & 42.128   & 0.856 & 0.642   & 3.27     & 0.673 & 0.288   \\
P36   & 4.428    & 0.714 & 0.619   & 5.195    & 0.776 & 0.472   & 4        & 0.645 & 0.442   \\
P364  & 4.541    & 0.709 & 0.553   & 5.545    & 0.707 & 0.481   & 3.47     & 0.69  & 0.478   \\
P37   & 4.876    & 0.718 & 0.477   & 5.863    & 0.787 & 0.489   & 0.7      & 0.56  & 0.447   \\
P39   & 5.230    & 0.722 & 0.413   & 6.051    & 0.683 & 0.450   & 0.6      & 0.59  & 0.243   \\
P407  & 5.456    & 0.730 & 0.611   & 6.149    & 0.680 & 0.483   & 0.04     & 0.358 & 0.280   \\
P413  & 5.632    & 0.809 & 0.539   & 6.346    & 0.740 & 0.522   & 0.9      & 0.63  & 0.252   \\
P449  & 6.005    & 0.710 & 0.717   & 7.150    & 0.743 & 0.501   & 0.6      & 0.53  & 0.373   \\
P463  & 6.088    & 0.776 & 0.467   & 7.697    & 0.735 & 0.437   & 1.7      & 0.65  & 0.489   \\
P47   & 6.356    & 0.776 & 0.739   & 8.078    & 0.701 & 0.524   & 11.7     & 0.8   & 0.730   \\
P495  & 6.455    & 0.709 & 0.513   & 8.209    & 0.744 & 0.569   & 1.0      & 0.505 & 0.585   \\
P530  & 7.280    & 0.785 & 0.688   & 8.598    & 0.765 & 0.558   & 1.84     & 0.552 & 0.615   \\
P740  & 7.324    & 0.769 & 0.518   & 8.840    & 0.737 & 0.477   & 2.1      & 0.505 & 0.796   \\
P937  & 8.623    & 0.725 & 0.572   & 9.594    & 0.782 & 0.621   & 2.53     & 0.673 & 0.752   \\ \bottomrule
\end{tabular}
\label{tab:levels_of_localisation_pararel}
\end{table}

\newpage
\section{Causal Tracing}
\label{app:causal_tracing}

In this section we present an extended analysis of our causal tracing analysis.
We use the same hyperparameters and settings as suggested by 
\citet{mengLocatingEditingFactual2022}.

\cites{mengLocatingEditingFactual2022} causal tracing experiment is conducted in
three steps:
\begin{enumerate}
    \item In the clean run, they pass a prompt into the model and record all the
    hidden activation values.
    \item Then, they conduct a corrupt run.  They perturb the prompt's subject
    by adding a noise value $\epsilon$ to the input embedding.  Because of the
    obfuscation, the model will like generate an incorrect answer.
    \item Finally, in the corrupted-with-restoration run, they let the model run
    with the corrupted value.  But every time for a pair of token and layer,
    they replace the corrupted output with the original clean state.  Then they
    let the model continue without further intervention.  If this restoration
    can attenuate the effect of obfuscation, they interpret this state as
    having a strong causal importance.  The difference between the corrupted
    output probability and the output probability after restoring one location
    is called the {\it indirect effect} of that location.
\end{enumerate}

\begin{figure}
\centering

\begin{subfigure}[b]{\linewidth}
\hfill\includegraphics[width=\linewidth]{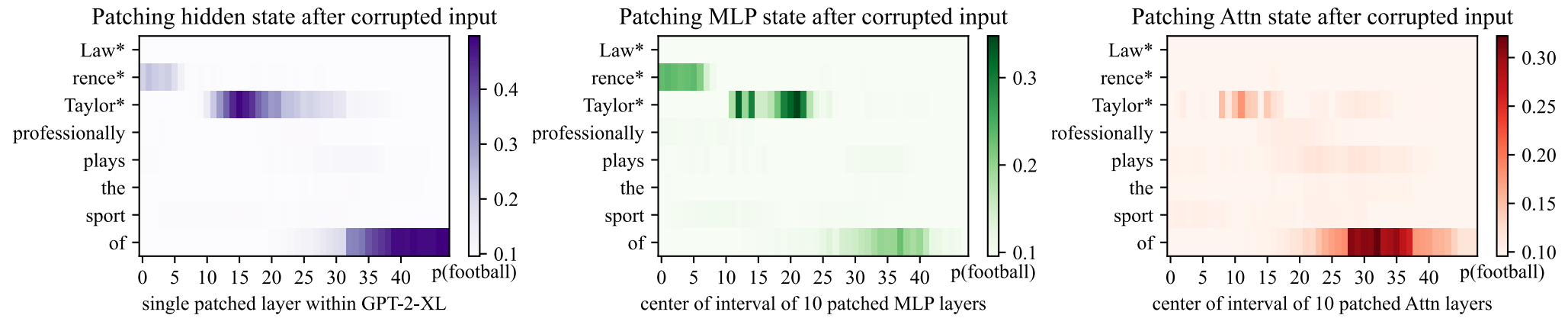}
\caption{Many factual causal traces also do not show this distinction.  Example
taken from \cites{mengLocatingEditingFactual2022} Figure 10c.  The MLP module
show causality at both the early and the late site.}
\vspace{1em}
\end{subfigure}

\begin{subfigure}[b]{\linewidth}
\includegraphics[width=0.33\linewidth]{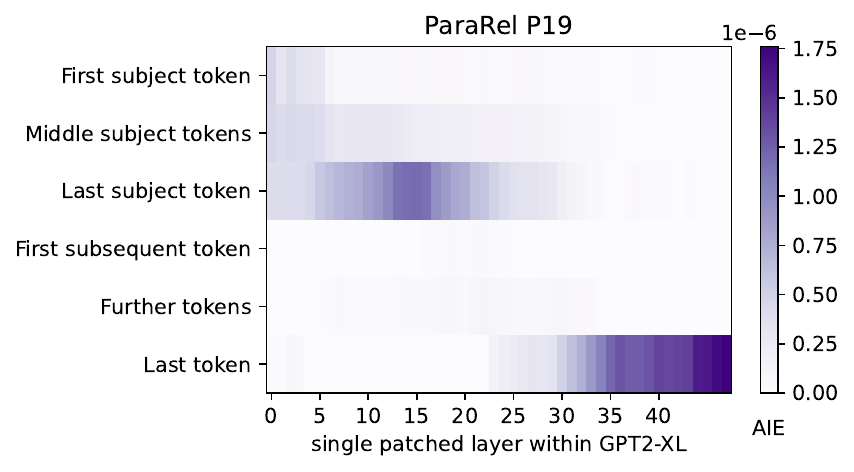}
\includegraphics[width=0.33\linewidth]{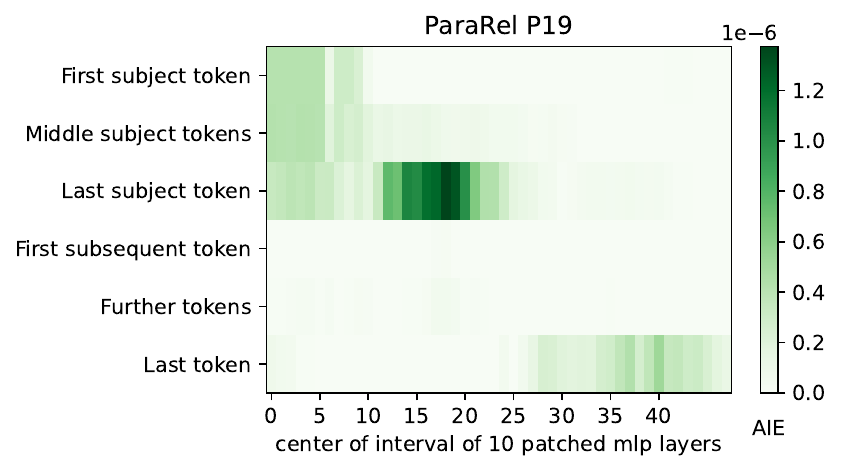}
\includegraphics[width=0.33\linewidth]{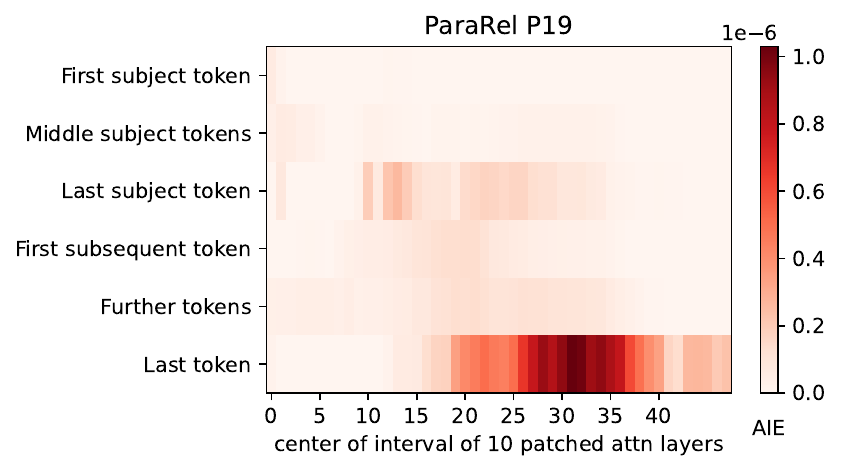}
\caption{We reproduced \cites{mengLocatingEditingFactual2022} calculation of the
average indirect effect of individual model components.  We do see the
separation between MLP and attention modules.  However, we can also see that
there is a weaker but discernible MLP causality at the late site.  This
shows that the previous example is not a negligible anomaly.  The causal
tracing pattern is less stable than \citet{mengLocatingEditingFactual2022}
originally conjectured.}
\vspace{1em}
\end{subfigure}

\begin{subfigure}[b]{\linewidth}
\includegraphics[width=0.33\linewidth]{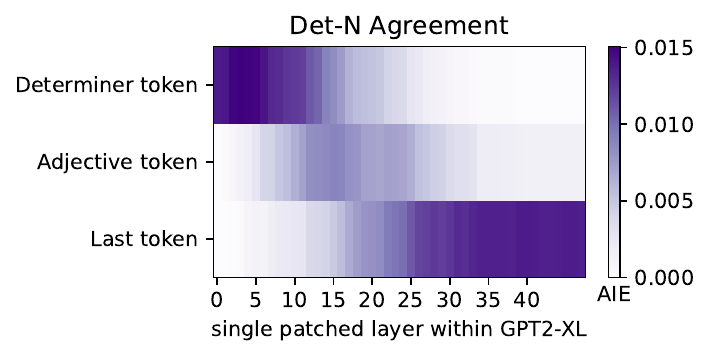}
\includegraphics[width=0.33\linewidth]{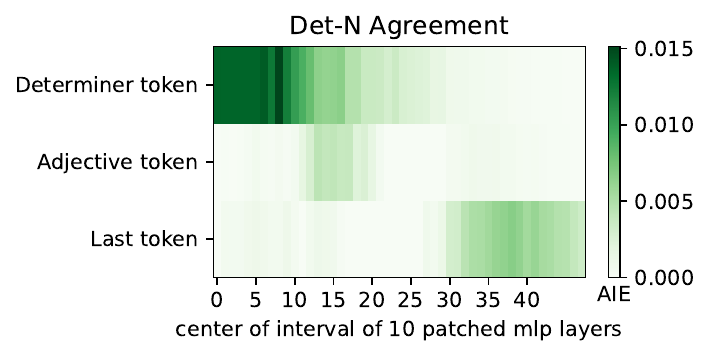}
\includegraphics[width=0.33\linewidth]{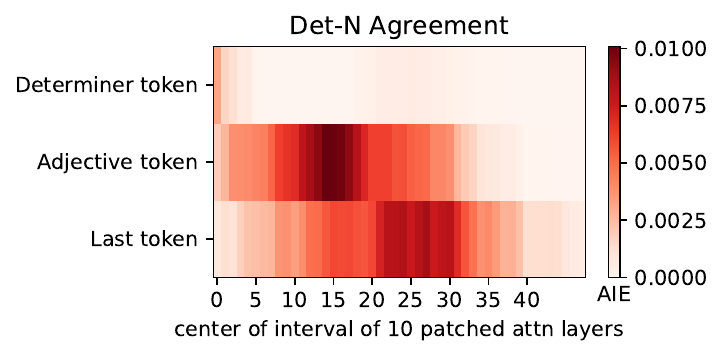}
\caption{The pattern that the MLP modules occupy an early site and attention
modules occupy a late site do not persist for determiner-noun agreement.  We
used the determiner\_noun\_agreement\_with\_adjective\_1 paradigm for this
experiment as it contains the adjective token that is analogous to the first
subsequent and further tokens that \citet{mengLocatingEditingFactual2022}
investigated.}
\vspace{1em}
\end{subfigure}

\begin{subfigure}[b]{\linewidth}
\includegraphics[width=0.33\linewidth]{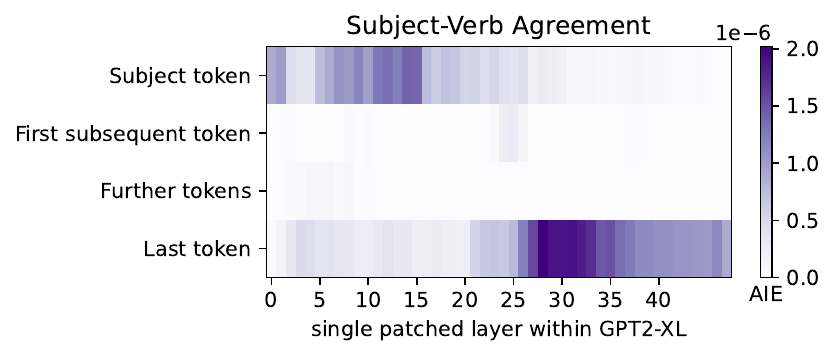}
\includegraphics[width=0.33\linewidth]{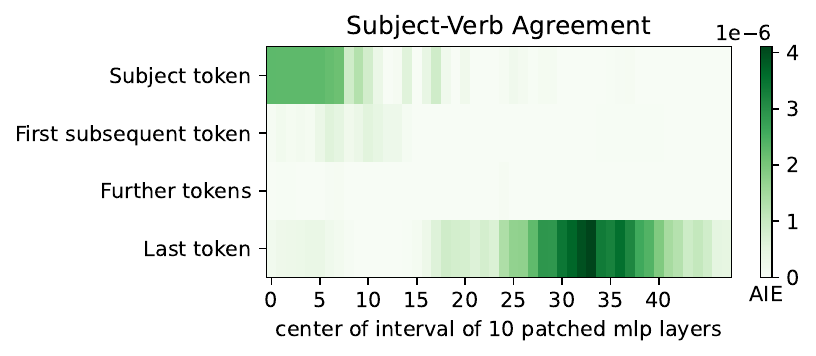}
\includegraphics[width=0.33\linewidth]{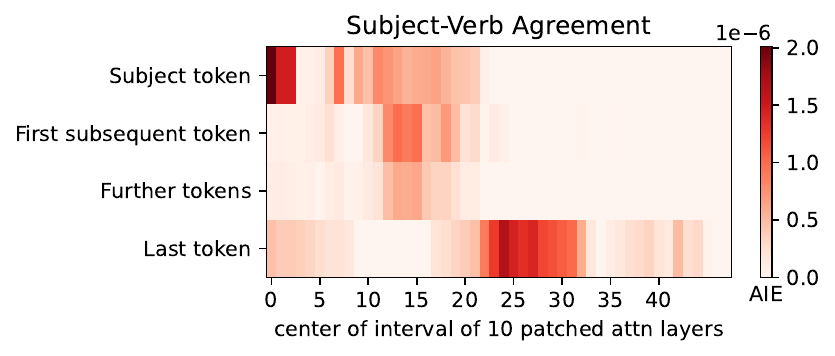}
\caption{The pattern of two distinct early and late sites is less apparent for
subject-verb agreement.  In fact, the MLP modules have the strongest causality
at higher layers (30-35) than attention ($\sim$25).}
\vspace{1em}
\end{subfigure}

\caption{Average indirect effect of different model component over multiple
samples.}
\label{fig:ie}
\end{figure}

However, all of the analysis are based on observing individual causal traces.
\citet{mengLocatingEditingFactual2022} also proposed to compute the average 
indirect effect over larger quantity of sentences.  In Table \ref{fig:ie} we
present our analysis of average indirect effects across different types of
information.  The result confirms our findings in Section \ref{sec:rome}.

\newpage

\section{Evaluation of Symmetry and Synonym}
\label{sec:extra_evaluation}

\begin{table}
\centering\scriptsize
\caption{The construction of our evaluation corpus for relation symmetry and synonym replacement}
\begin{tabular}{l|ll|l} \toprule
Relation & \multicolumn{2}{c|}{Edit Prompt} & Evaluate Prompt \\ \midrule
Symmetry & Template: & [S] is the capital of [T$\rightarrow$T$^*$].                         & The capital of [T$^*$] is [S$^*\rightarrow$S].           \\
P1376    & Example:  & Ottawa is the capital of Canada$\rightarrow$Italy.                   & The capital of Italy is Rome$\rightarrow$Ottawa.         \\ \midrule
Symmetry & Template: & The capital of [S] is [T$\rightarrow$T$^*$].                         & [T$^*$] is the capital of [S$^*\rightarrow$S].           \\
P36      & Example:  & The capital of Canada is Ottawa$\rightarrow$Rome.                    & Rome is the capital of Italy$\rightarrow$Canada.         \\ \midrule
Synonym  & Template: & [S] works in the field of [T$\rightarrow$T$^*$].                     & [S] is a [T$_s\rightarrow$T$_s^*$]                       \\
P101     & Example:  & Anaxagoras works in the field of philosophy$\rightarrow$linguistics. & Anaxagoras is a famous philosopher$\rightarrow$linguist. \\
\bottomrule
\end{tabular}
\label{tab:eval_overview}
\end{table}

We construct the datasets used for both the symmetry and synonym evaluation from
{\sc ParaRel} relations.  Table \ref{tab:eval_overview} shows an overview of the
data construction process.

\paragraph{Symmetry}
We used the two bijective relations: P1376 (capital of) and P36 (capital) to
evaluate the symmetry property.

For each P1376 (capital of) relation $(s,t,r)$, we edit the model using the
prompt ``[S] is the capital of'' and change the target from $t$ to $t^*$.  $t^*$
is another city
from the corpus.  Then, we prompt the model with ``The capital of [T$^*$] is''
and see if the model outputs $s$ with a higher probability than the original
$s^*$.  We identified 234 relations in total.

Similarly, for relation P36 (capital), we edit the model using the original
prompt ``The capital of [S] is'' and try to change the target from $t$ to $t^*$.
$t^*$ is another country/state from the corpus.  Again, if the model outputs $s$
with a higher probability than the original $s^*$, we count the evaluation as
success.  We identified 703 relations.

\paragraph{Synonym}

For synonym replacement, we use P101 (field of work).
We first rewrite each field of work to the occupation name.  For example, {\it
linguistics} $\rightarrow$ {\it linguist} and {\it aviation} $\rightarrow$ {\it
pilot}.

Through the process, we identified several mistakes in the original P101 data. 
For example, some of the field of work targets are already names of occupation,
resulting in ill-formed prompts such as ``Clyde Tombaugh works in the field of 
{\bf astronomer}.''  Some of the fields of work, typically country names, cannot
be converted into occupation names.  For example ``Mark Mazower works in the
field of {\bf Balkans}.''  We discard data entries with these issues and we
collect 50 distinct field of works with their occupations.  Finally, we obtain
568 entries.

We edit the model with the original prompt ``[S] works in the filed of'' and
edit the target from $t$ to $t^*$.  $t^*$ is another field of work we collected
from P101.  Then, we use the prompt ``[S] is a'' to elicit response from the
edited model.  Let $t_s$ and $t_s^*$ be the two occupation names correspond to
$t$ and $t^*$, we want to see if the model can also assign a higher probability
to the new target synonym $t_s^*$ than the original $t_s$.

For example, first we edit the model from ``Anaxagoras works in the field of
\underline{philosophy}'' to \underline{linguistics}.  Then, we prompt the model
with ``Anaxagoras is a famous'' an see whether the model assign a higher probability to
{\it linguist} rather than {\it philosopher}.

\end{document}